\documentclass[10pt,twocolumn,letterpaper]{article}

\usepackage[pagenumbers]{cvpr}   
\makeatletter
\@namedef{ver@everyshi.sty}{}
\makeatother
\usepackage[table]{xcolor}
\usepackage{graphicx}
\usepackage{amsmath}
\usepackage{amssymb}
\usepackage{booktabs}
\usepackage{MnSymbol}
\usepackage{nicematrix}
\usepackage{diagbox}
\usepackage{multirow}
\usepackage{pbox}
\usepackage{booktabs}
\usepackage{makecell}
\usepackage{hhline}
\usepackage{algorithm} 
\usepackage{algpseudocode} 
\usepackage{multirow}
\definecolor{mygray}{gray}{0.95}
\usepackage{stfloats}

\usepackage[pagebackref,breaklinks,colorlinks]{hyperref}

\newlength\savedwidth
\newcommand\whline{\noalign{\global\savedwidth\arrayrulewidth\global\arrayrulewidth 0.8pt}\hline\noalign{\global\arrayrulewidth\savedwidth}}

\usepackage[capitalize]{cleveref}
\crefname{section}{Sec.}{Secs.}
\Crefname{section}{Section}{Sections}
\Crefname{table}{Table}{Tables}
\crefname{table}{Tab.}{Tabs.}

\def\cvprPaperID{6282} 
\def\confName{CVPR}
\def\confYear{2023}

\begin{document}

\title{Scale-free and Task-agnostic Attack: Generating Photo-realistic Adversarial Patterns with Patch Quilting Generator}

\author{Xiangbo Gao\\
University of California, Irvine\\
{\tt\small xiangbog@uci.edu}
\and
Cheng Luo\\
Shenzhen University\\
{\tt\small luocheng2020@email.szu.edu.cn}
\and
Qinliang Lin\\
Shenzhen University\\
{\tt\small 2017192020@email.szu.edu.cn}
\and
Weicheng Xie\thanks{Corresponding author.}\\
Shenzhen University\\
{\tt\small wcxie@szu.edu.cn}
\and
Minmin Liu\\
Shenzhen University\\
{\tt\small liuminmin2020@email.szu.edu.cn}
\and
Linlin Shen\\
Shenzhen University\\
{\tt\small llshen@szu.edu.cn}
\and
Keerthy Kusumam\\
University of Nottingham\\
{\tt\small keerthy.kusumam2@nottingham.ac.uk}
\and
Siyang Song\thanks{Corresponding author.}\\
University of Cambridge\\
{\tt\small ss2796@cam.ac.uk}
}


\twocolumn[{\maketitle
\vspace{-0.2cm}
\begin{figure}[H]
\setlength{\linewidth}{\textwidth}
\setlength{\hsize}{\textwidth}
\centering
\includegraphics[width=1.85\columnwidth]{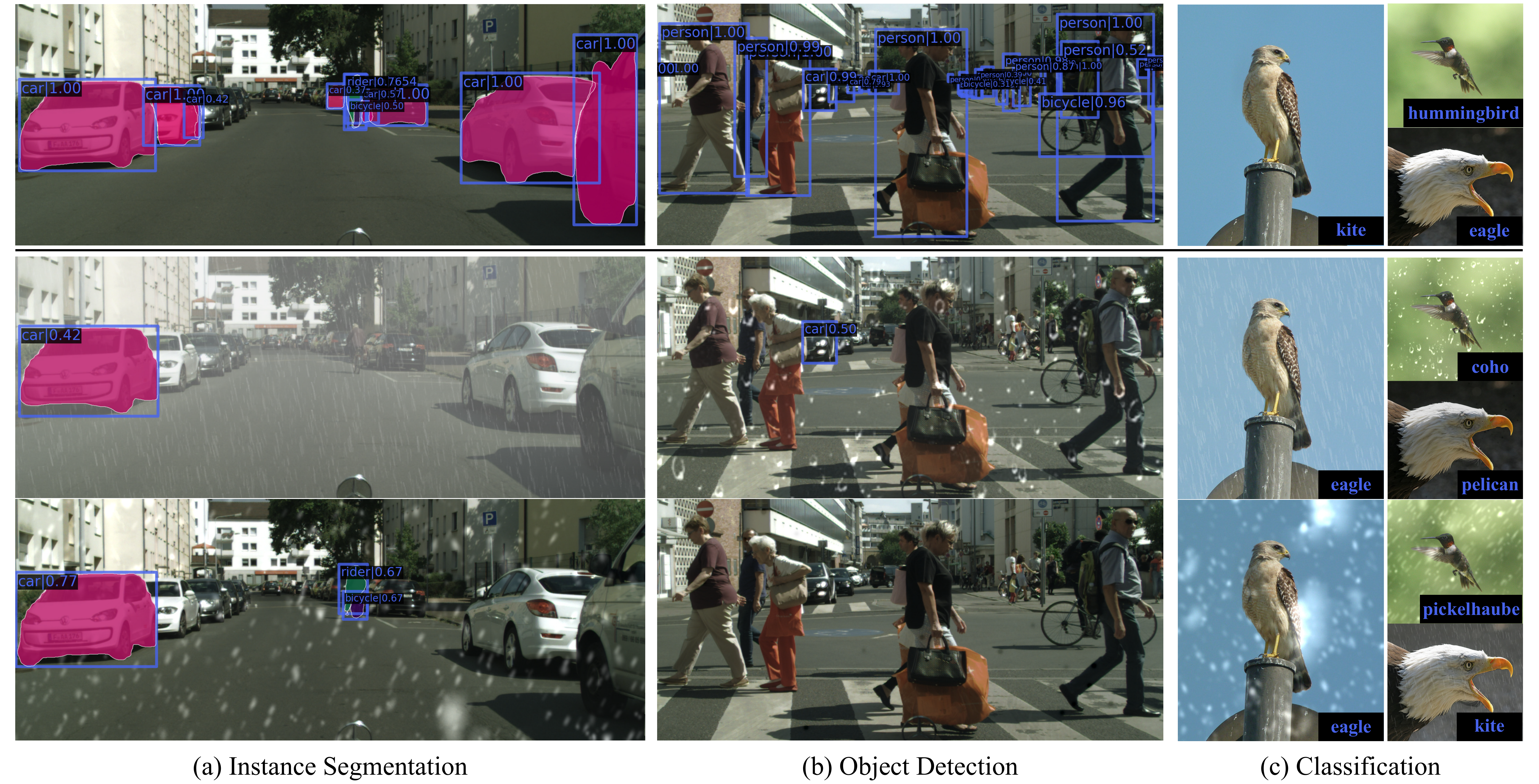}
\caption{Our approach can \textbf{attack various vision tasks}, \emph{i.e,} (a) instance segmentation, (b) object detection and (c) classification with \textbf{images of arbitrary scales}. The first row shows benign examples and the last two rows display images attacked by our approach.}
\label{img_intro}
\end{figure}}]

\begin{abstract}
\noindent Traditional $L_p$ norm-restricted image attack algorithms suffer from poor transferability to black box scenarios and poor robustness to defense algorithms. Recent CNN generator-based attack approaches can synthesize unrestricted and semantically meaningful entities to the image, which is shown to be transferable and robust. However, such methods attack images by either synthesizing local adversarial entities, which are only suitable for attacking specific contents or performing global attacks, which are only applicable to a specific image scale. In this paper, we propose a novel Patch Quilting Generative Adversarial Networks (PQ-GAN) to learn the first scale-free CNN generator that can be applied to attack images with arbitrary scales for various computer vision tasks. The principal investigation on transferability of the generated adversarial examples, robustness to defense frameworks, and visual quality assessment show that the proposed PQG-based attack framework outperforms the other nine state-of-the-art adversarial attack approaches when attacking the neural networks trained on two standard evaluation datasets (i.e., ImageNet and CityScapes). Our anonymous code is made available at \url{https://anonymous.4open.science/r/PQAttack-0781}.
\end{abstract}

\section{Introduction}
\label{sec:intro}

\begin{table*}[htbp]
\renewcommand{\arraystretch}{0.85}
\small
\centering
\begin{tabular}{ccccccc}
\hline
Work            &Photo-realistic                                  & Scale-free    & Task-agnostic     & Black-box     & Defense \\ \hline
UAP$_{\ \text{CVPR'2018}}$\cite{mopuri2018generalizable} &                      & $\checkmark$  & $\checkmark$      &             & $\checkmark$      \\
AdvFaces$_{\ \text{IJCB'2020}}$\cite{deb2020advfaces} &  $\checkmark$                           &   &                   & $\checkmark$            &       \\
ColorFool$_{\ \text{CVPR'2020}}$\cite{shamsabadi2020colorfool} & $\checkmark$   &$\checkmark$&                   & $\checkmark$           & $\checkmark$ \\
RA-AVA$_{\ \text{IJCAI'2021}}$\cite{tian2021ava} & $\checkmark$   & $\checkmark$  &                   &          & $\checkmark$ \\
Shadows Attack$_{\ \text{CVPR'2022}}$\cite{zhong2022shadows}  & $\checkmark$    &   $\checkmark$            &                   & $\checkmark$           & $\checkmark$     \\
Ours                      & $\checkmark$                        & $\checkmark$  & $\checkmark$      & $\checkmark$           & $\checkmark$ \\
\hline
\end{tabular}
\caption{The comparison of our attack method to the previous works. Our attack method is scale-free, task-agnostic, and imperceptible. Meanwhile, we experimentally show that our method has high black-box transferability and is robust to existing defense algorithms.}
\label{tb:comparsion-intro}
\vspace{-0.2cm}
\end{table*}


Deep Neural Networks (DNNs) are vulnerable to adversarial examples, \emph{e.g.}, images with carefully designed adversarial perturbations can easily mislead well-trained DNNs to output incorrect predictions. To overcome such malicious attacks, several adversarial defense algorithms have been proposed, which, in turns, simulate the development of robust adversarial attack algorithms to disrupt these defenses. Therefore, investigating robust and powerful image attack algorithms plays a crucial role in progressing current research toward developing strong defense algorithms.
\par Traditional image attack approaches \cite{goodfellow2014explaining, carlini2017towards, dong2018boosting, wiyatno2018maximal, moosavi2016deepfool, feinman2017detecting, madry2017towards, su2019one} focus on generating perturbations at the pixel-level with $L_p$-norm restrictions, which possess strong capabilities to mislead predictors but are imperceptible to human eyes. However, the attack strength of such restricted approaches cannot be transferred to unseen networks and the noises are easily defensed \cite{zhang2019adversarial,das2017keeping}. Subsequently, many studies devote their efforts to formulating methods that deliver more robust attack patterns against the defense algorithms. Some of them add noisy perturbation with weaker restriction \cite{mopuri2018generalizable,sun2022towards} or even without restriction \cite{chen2021unrestricted}. Despite that larger perturbation boosts up the transferability and robustness of attack algorithms, the perturbation is perceptible to human eyes and the adversarial examples not photo-realistic. 





To solve this problem, some studies employ generative adversarial networks (GAN)\cite{goodfellow2014generative} to generate semantically meaningful local entities and synthesize them to the image\cite{sharif2016accessorize,komkov2021advhat,yin2021adv,hu2022protecting,zhong2022shadows,kong2020physgan} or to change the texture of a particular area of the image\cite{duan2020adversarial,hu2022adversarial}. Such GAN-style methods can generate robust and transferable adversarial examples with high image quality. However, such methods are designed for a specific task, such as face recognition \cite{komkov2021advhat} and vehicle motion prediction \cite{kong2020physgan}. (\textbf{Problem 1}). Alternatively, instead of generating a local entity, some works propose methods that generate adversarial examples in an end-to-end manner where a global adversarial perturbation is carefully hidden in the target image \cite{liu2019rob,jandial2019advgan++,deb2020advfaces,bai2021ai,xiao2018generating, poursaeed2018generative, zhang2020generalizing, naseer2021generating}. However, due to the limitation of traditional GAN structure, these methods can only generate adversarial examples of one particular or highly limited scale (\textbf{Problem 2}). The whole generative network must be re-trained when changing the target image scale. It is worth mentioning that some works can generate adversarial examples with global semantic patterns without using GAN \cite{niu2021morie, bhattad2019unrestricted, shamsabadi2020colorfool, gao2021advhaze, tian2021ava}. However, these methods can only generate a specific attack pattern with carefully designed math formulation, which cannot be extended to general usage.

In this paper, we propose a novel Patch Quilting Generative Adversarial Network (PQ-GAN) to address the aforementioned problems of existing image attack algorithms. The PQ-GAN learns three cascaded generators that can synthesize photo-realistic, scale-free patterns to attack target images of any scale on the whole-image level (globally) (\textbf{addressing the problem 2}). Importantly, the synthesized realistic and semantically meaningful pattern ensures the visual quality of the adversarial examples. Task agnostic property allows our approach to be applied to generating various photo-realistic patterns, e.g., rain streaks, snow flakes, and camera lens dirt. It can be applied to many computer vision tasks such as image classification, object detection, instance segmentation, etc. (\textbf{addressing the problem 1}). In addition, our approach generates patterns with unrestricted pixel value, ensuring transferability and robustness. The main advantages of our approach compared to existing approaches are listed in Table~\ref{tb:comparsion-intro}. The main contributions and novelties of the proposed approach are summarized as follows:
\begin{itemize}
    \item We propose a PQ-GAN-based unrestricted adversarial attack pipeline that generates various global adversarial patterns to attack images. This method is not limited to attack images of any particular scale or computer vision task, namely, scale-free and task-agnostic.
        
    \item We propose a novel PQ-GAN which can learn three cascaded generators that synthesize photo-realistic and semantically meaningful images of any scale without any distortion or discontinuity. To the best of our knowledge, this is the first deep learning generator that can synthesize images of any scale. 
    
    \item The principal investigation results demonstrate that our approach delivers state-of-the-art attack strength and transferability among the existing synthesis-based attack methods, and our experiments verify the dominance of the proposed approach against various types of defense algorithms.
    
\end{itemize}

\begin{figure*}[t]
\centering
\includegraphics[width=1.84\columnwidth]{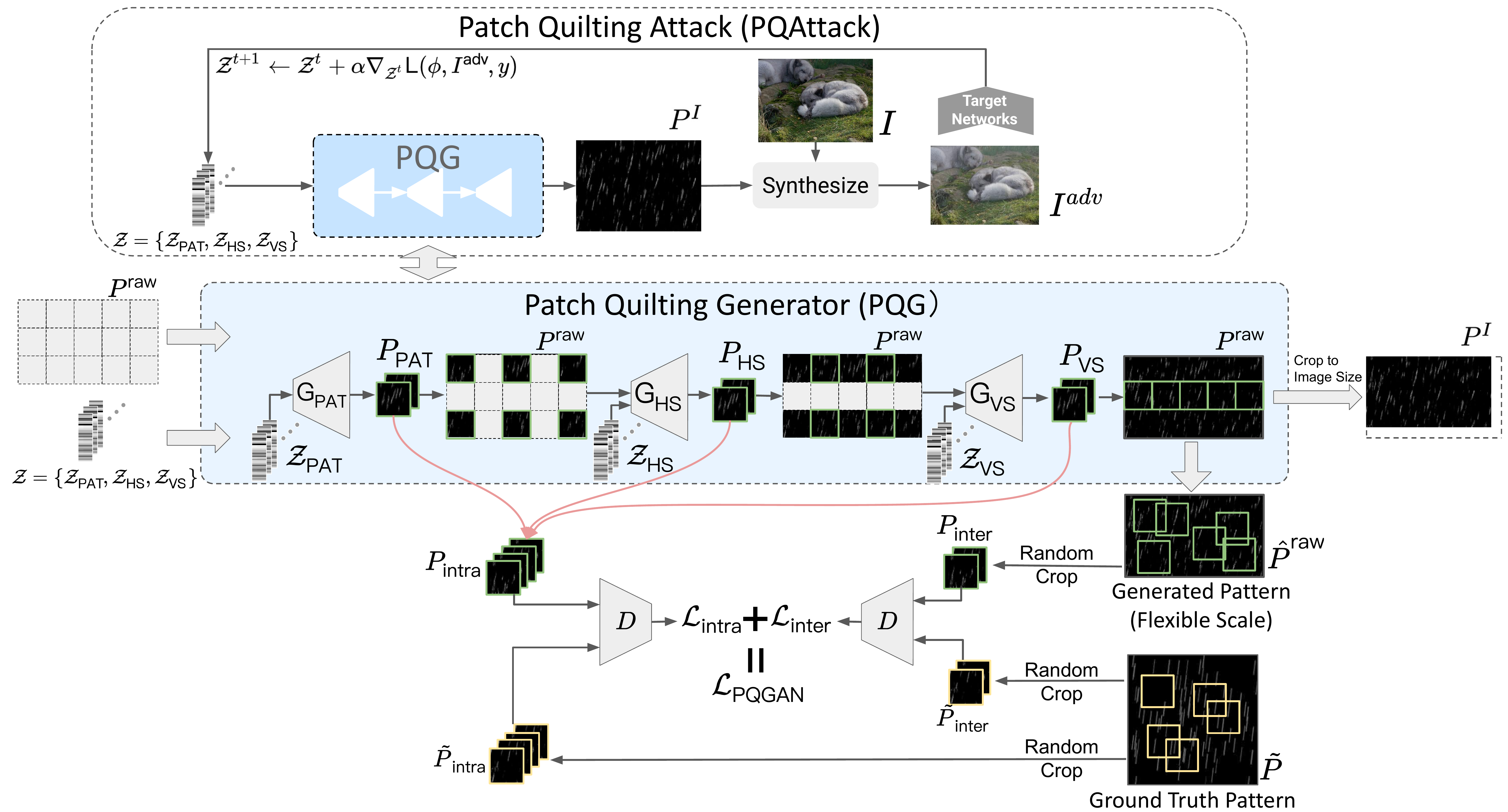} 
    \caption{Illustration of the Patch Quilting Attack pipeline (Top) and the training strategy of the Patch Quilting Generator (Bottom).}
\label{PQGAN_flow}
\vspace{-0.2cm}
\end{figure*}

\section{Related Work}

\noindent \textbf{Task/contents-specific Adversarial Attack:}\quad Traditional adversarial attack strategies \cite{goodfellow2014explaining,carlini2017towards,dong2018boosting,dong2020greedyfool,su2019one} are extensively researched to generate adversarial examples by adding $L_{p}$ bounded adversarial perturbations to the target images. Such strategies usually lack robustness to the defense algorithms \cite{das2018shield, zhang2019adversarial, liao2018defense, meng2017magnet, shen2017ape, samangouei2018defense} as the highly restricted perturbations are easily removed. As a result, some recent studies attempted to replace the $L_p$ constraint with either perceptual similarity \cite{wang2019invisible} or JND$_p$ \cite{duan2020adversarial} constraint. Recently, Luo \emph{et al}. \cite{luo2022frequency} further propose to add perturbations to attack images based on semantic similarity. To further improve the robustness, others \cite{liu2018dpatch,lee2019physical,yang2020patchattack} propose to generate a semantic meaningful local patch to attack images. For example, some studies attack facial images by adding a glass \cite{sharif2016accessorize}, a hat \cite{komkov2021advhat}, or makeups \cite{yin2021adv,hu2022protecting} to the target face. Eykholt \emph{et al}. \cite{eykholt2018robust} propose to add graffiti to attack road signs. Duan \emph{et al}. \cite{duan2020adversarial} and Zhong \emph{et al}. \cite{zhong2022shadows} perform style transfer or add shadow to attack a selected region in the image. Besides, some studies are built on specific math formulas to generate specific robust and global attack patterns, including haze \cite{gao2021advhaze}, vignetting \cite{tian2021ava}, and moire pattern\cite{niu2021morie}. While the aforementioned strategies can generate robust attacks, most of them still suffer from three main problems: (i) some of them can be only applied to limited application scenarios such as the human face \cite{sharif2016accessorize,komkov2021advhat,yin2021adv,hu2022protecting} or road sign \cite{eykholt2018robust}; (ii) most of them are only suitable for attacking networks of a specific computer vision task (e.g., image classification networks \cite{brown2017adversarial,yang2020patchattack} or object detection networks \cite{liu2018dpatch,lee2019physical}); and (iii) they are still not robust enough to recently proposed advanced defense algorithms \cite{das2018shield, zhang2019adversarial, liao2018defense, meng2017magnet, shen2017ape}.

\noindent \textbf{Scale-specific Generative Model-based Adversarial Attack:}\quad To obtain human-realistic and more robust attack patterns with high diversity for various scenarios, recent approaches frequently employ Generative Adversarial Networks (GAN)-style generators to synthesize semantic patterns\cite{liu2019rob,bai2021ai, komkov2021advhat, yin2021adv}. This is because GAN \cite{goodfellow2014generative} has been widely used in many areas due to its capability of learning and generating various robust data distributions. However, these approaches can only generate attack patterns of a fixed scale decided by the pre-defined generator because the traditional GAN structure is designed for fix-scale image generation. Although some researchers propose hierarchical \cite{shaham2019singan,karnewar2020msg,chen2021hierarchical} or growing convolution \cite{karras2017progressive} architectures to generate images of different resolutions by picking the feature map from different layers, they can only generate patterns of a small set of pre-defined resolutions, which cannot be fully reusable under scale-agnostic adversarial attack scenario. In this paper, we propose an image Patch Quilting Generative Adversarial Network (PQ-GAN) that can generate patterns of any scale without retraining the model.

\section{Methodology}

\subsection{Patch Quiliting Attack}

Given a target network of agnostic image analysis task with model parameters $\phi$ and a loss function $\mathcal{L}(\phi, I, y)$ used for model training, where $y$ is the label of the benign image $I$, the goal to find an adversarial example $I^{\text{adv}}$ that maximizes $\mathcal{L}(\phi, I^{\text{adv}}, y)$ under the restriction that $I^{\text{adv}}$ is perceptually natural. In particular, Fig.~\ref{PQGAN_flow} illustrates our scale-agnostic generative model (whose model parameters are represented as $\psi$), namely Patch Quilting Generator (PQG). The PQG takes a set of latent embeddings $\mathcal{Z}$ initialized with Gaussian distribution of mean 0 and standard deviation 1 as the input, which controls the characteristic of the pattern, and outputs a photo-realistic pattern $P^I$. Then, $P^I$ is synthesized to the target image $I$ to produce an adversarial example $I^{\text{adv}}$ that is perceptually natural. Now the problem is transformed to find a set of latent embeddings $\mathcal{Z}$ that maximize $\mathcal{L}(\phi, I^{\text{adv}}, y)$,  which is formulated as:
\begin{equation}
    \begin{split}
        \text{max}_{\mathcal{Z}}&\quad \mathcal{L}(\phi, I^{\text{adv}}, y) \\
        \textbf{\text{Subject to}}&\quad I^{\text{adv}} = \text{Syn}(I, P^I) \\
        \text{where} &\quad P^I = \text{PQG}(\mathcal{Z}, \psi)
    \end{split}
\end{equation}
Notice that PQG is a pre-trained model that does not need to be retrained in this attack stage. We explain the latent embeddings $\mathcal{Z}$ and how the PQG is designed to be scale-agnostic in detail in Sec. \ref{Sec3d2}. $\text{Syn}(\cdot)$ is a customized synthesis function depends on the target pattern, (\emph{i.e.} pixel-wised addition, or depth-aware synthesis \cite{hu2019depth}).

To achieve the adversarial objective, we simply apply gradient ascent to the loss function being use for model train to update the latent embeddings $\mathcal{Z}$ through iterations, \emph{i.e.},
\begin{equation}
\label{atk_iter}
    \begin{split}
        & \mathcal{Z}^{t+1} \leftarrow {\mathcal{Z}^t} + \alpha\nabla_{\mathcal{Z}^t}\mathcal{L}(\phi, I^{\text{adv}}, y)
    \end{split}
\end{equation}
where $t$ is the time-stamp and $\alpha$ denotes the learning rate. We put $\mathcal{Z}$ of the last iteration into PQG to generate the adversarial example.

\subsection{Patch Quilting GAN}\label{Sec3d2} 
\subsubsection{Scale-free Image Generation via PQG}

The Patch Quilting Generator ($\text{PQG}$) consists of three cascaded generators $\text{G}_\text{PAT}$, $\text{G}_\text{HS}$, and $\text{G}_\text{VS}$ with the learnable weights $\psi_\text{PAT}, \psi_\text{HS}, \psi_\text{VS}$, where each can generate image patches of the scale $h\times w$. 
PQG takes three sets of latent embeddings $\mathcal{Z}=\{\mathcal{Z}_\text{PAT},\mathcal{Z}_\text{HS}, \mathcal{Z}_\text{VS}\}$ as the input, and outputs a set of patches of the scale $h\times w$, which contain the required attack pattern. These patches can then be combined as a smooth and photo-realistic global attack pattern $P^{I}$ whose scale can be customized based on the target image.

As shown in Fig.~\ref{PQGAN_flow}, given a target image $I$ with the scale of $H \times W$, our PQG first initializes an attack pattern $P^{\text{raw}} \in \mathbb{R}^{\hat{H} \times \hat{W}}$, which consists of a integral number of empty patches of size $h\times w$, \emph{i.e.} $\hat{H}, \hat{W}$ are formulated as:
\begin{equation}
    \hat{H} = N_h \times h, \quad   \hat{W} = N_w \times w
\end{equation}
where $N_h=\text{ceil}(\frac{H}{h})$, $N_w=\text{ceil}(\frac{W}{w})$ denote the minimum number of patches that are required to fill up each row and column, $\text{ceil}(\cdot)$ means rounding up to an integer.
Then three generators then generates a set of attack patches as follows:

    Firstly, $\text{G}_{\text{PAT}}$ takes a set of latent embeddings $\mathcal{Z}_{\text{PAT}}$ to generate a set of attack patches $P_\text{PAT}$ to fill up non-adjacent odd rows and columns in the $P^{\text{raw}}$. Let $N^\text{PAT}_h=\text{ceil}(\frac{N_h}{2})$ and $N^\text{PAT}_w=\text{ceil}(\frac{N_w}{2})$.  This step can be formulated as:
    \begin{equation}
    \small
    \label{eqPAT}
    \begin{split}
        \quad & p^{\text{raw}}_{2a-1,2b-1}\in P_{\text{PAT}} = \text{G}_\text{PAT}(\mathcal{Z_\text{PAT}}, \psi_{\text{PAT}})\\
        \textbf{\text{Subject to}}\quad 
        & \mathcal{Z}_{\text{PAT}} = \{z_{2a-1,2b-1} \in \mathcal{N}(0,1)^k\}\\
        & a \in \{1,2,\cdots,N^\text{PAT}_h\},\ b \in \{1,2,\cdots,N^\text{PAT}_w\}\\
    \end{split}
    \end{equation}
    where $p^{\text{raw}}_{2a-1,2b-1}$ denotes the image patch located at the $2a-1_\text{th}$ row and $2b-1_\text{th}$, while $z_{2a-1,2b-1} \in \mathcal{N}(0,1)^k$ denotes the latent embedding of dimension $k$ being used to generate $p^{\text{raw}}_{2a-1,2b-1}$.
    
    Secondly, $\text{G}_\text{HS}$ generates a set of horizontal context-aware realistic adversarial attack patches $P_\text{HS}$ where each pathc fills up a horizontal gap in $P^{\text{raw}}$ based on not only $Z_\text{HS}$ but also its horizontal neighbours in $P^{\text{raw}}$, which are generated from $\text{G}_\text{PAT}$. By letting $N^\text{HS}_h=\text{ceil}(\frac{N_h}{2})$ and $N^\text{HS}_w=\text{ceil}(\frac{N_w}{2})-1$, this process is formulated as:
    
    \begin{equation}
    \small
    \label{eqHS}
    \begin{split}
        \quad & p^{\text{raw}}_{2a-1,2b}\in P_\text{HS} = \text{G}_\text{HS}(\mathcal{Z}_\text{HS}, p^{\text{raw}}_{2a\pm 1,2b+1}, \psi_{\text{HS}})\\
        \textbf{\text{Subject to}}\quad 
        & \mathcal{Z}_{\text{HS}} = \{z_{2a-1,2b} \in \mathcal{N}(0,1)^k\}\\
        & a \in \{1,2,\cdots,N^\text{HS}_h\},\ b \in \{1,2,\cdots,N^\text{HS}_w\}\\
    \end{split}
    \end{equation}
    Notice that $p^{\text{raw}}_{2a-1,2b\pm 1} \in P_{\text{PAT}}$.
    
    Finally, $\text{G}_\text{VS}$ generates a set of vertical context-aware realistic adversarial attack patches $P_\text{VS}$, targeting on filling up the rest regions (all vertical gaps) in $P^{\text{raw}}$. Specifically, each patch generated by  $\text{G}_\text{VS}$ fills up a gap based on not only $\mathcal{Z_\text{VS}}$ but also its vertical neighbours in $P^{\text{raw}}$. By letting $N^\text{VS}_h=\text{ceil}(\frac{N_h}{2})-1$ and $N^\text{VS}_w=N_w$,  which are produced from $\text{G}_\text{PAT}$ as: 
    \begin{equation}
    \small
    \label{eqVS}
    \begin{split}
        &p^{\text{raw}}_{2a,b}\in P_\text{VS} = \text{G}_\text{VS}(\mathcal{Z}_\text{VS}, p^{\text{raw}}_{2a\pm 1,b}, \psi_{\text{VS}})\\
        \textbf{\text{Subject to}}\quad 
        & \mathcal{Z}_{\text{VS}} = \{z_{2a,b} \in \mathcal{N}(0,1)^k\}\\
        & a \in \{1,2,\cdots,N^\text{VS}_h\},\ b \in \{1,2,\cdots,N^\text{VS}_w\}\\
    \end{split}
    \end{equation}
    Notice that $ p^{\text{raw}}_{2a\pm 1,b} \in P_{\text{PAT}}\bigcup P_{\text{HS}}$.
    
Consequently, a global pattern $\hat{P}^{\text{raw}}$ is obtained by filling all patches of the $P^{\text{raw}}$, where attack patches produced by three generators are concatenated.  We then remove the extra pixels of the $\hat{P}^{\text{raw}}  \in \mathbb{R}^{\hat{H} \times \hat{W}}$ to make it have the same size $H \times W$ to the target image $I$, which is denoted as the final $P^{I}$. In summary, the proposed PQG can not only synthesize global image attack patterns of any required scale without requiring re-training the network, but also allow the final produced pattern to be smooth, continuous and semantically meaningful.

\begin{figure*}[!htbp]
    \centering
    \includegraphics[width=2\columnwidth]{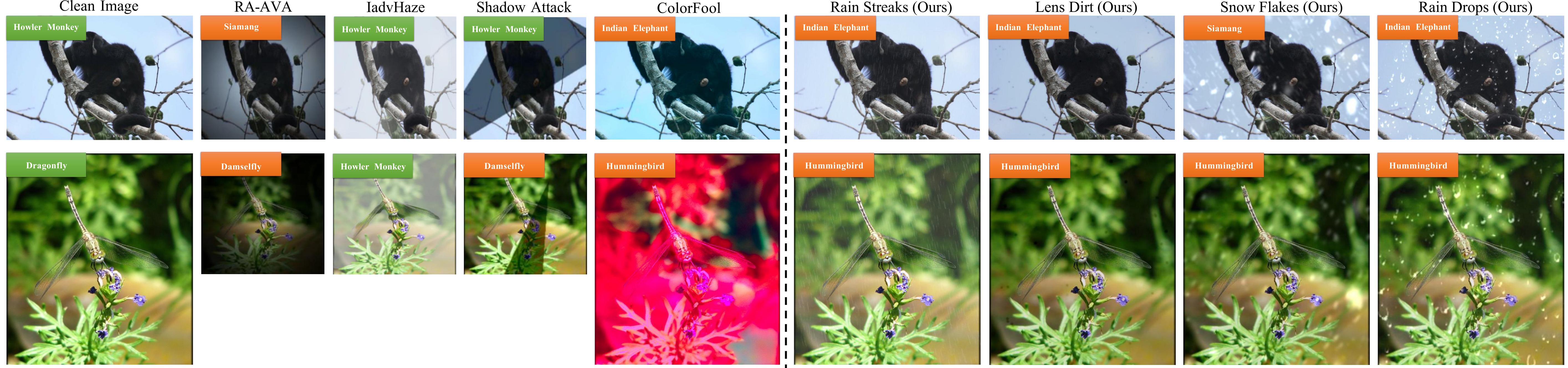}
    \caption{Visualization of adversarial examples generated by various attack methods. Our adversarial examples successfully mislead the target classifier. Note that our attack method can generate high-resolution adversarial examples with the original image scale.}
    \label{fig:comparison}
    \vspace{-0.2cm}
\end{figure*}

\subsubsection{PQ-GAN Optimization}

To learn three generators of the POG, we propose a GAN-style training strategy (called PQ-GAN). To produce a globally smooth and continuous attack pattern, we propose a \textbf{Intra-Patch Smoothness Loss} to ensure each generated attack patch to be smooth and photo-realistic, and a \textbf{Inter-Patch Smoothness Loss} to enforce the smoothness between neighbor patches.

\noindent\textbf{Intra-Patch Smoothness Loss:}\quad As displayed at the bottom of the Fig.~\ref{PQGAN_flow}, we collect all the patches $P_{\text{intra}} = P_\text{PAT} \bigcup P_\text{HS} \bigcup P_\text{VS}$ generated by generators of the POG, treating them as negative samples, while a set of positive samples $\tilde{P}_\text{intra} = \{\tilde{p}^m_\text{intra} |\ m = 1, 2, \cdots, M_\text{intra}\}$ are obtained by randomly cropping a set of ground truth patches of size $h\times w$ from the a ground truth Pattern $\tilde{P}$, where $M_\text{intra}$ equals the number of patches in $P_{\text{intra}}$. All negative and positive sample are then fed into a discriminator $\text{D}$ to calculate the generator loss and discriminator loss, respectively, based on the standard formulation of the Wasserstein GAN with gradient penalty\cite{arjovsky2017wasserstein}. This process is formulated as:
\small
\begin{align}
\mathcal{L}_{\text{intra}} = & \mathcal{L}_{\text{G}_\text{PAT}}(P_\text{intra}, \psi_\text{PAT}) +
    \mathcal{L}_{\text{G}_\text{HS}}(P_\text{intra}, \psi_\text{HS}) +  \mathcal{L}_{\text{G}_\text{VS}}(P_\text{intra}, \psi_\text{VS}) + \nonumber \\
    &\mathcal{L}_{\text{D}}(P_\text{intra}, \psi_\text{D})+ \mathcal{L}_{\text{D}}(\tilde{P}_\text{intra}, \psi_\text{D})
\end{align}
\normalsize

where $\mathcal{L}_{\text{G}_\text{PAT}}(P_\text{intra}, \psi_\text{PAT}),\ \mathcal{L}_{\text{G}_\text{HS}}(P_\text{intra}, \psi_\text{HS})$, and $\mathcal{L}_{\text{G}_\text{VS}}(P_\text{intra}, \psi_\text{VS})$ denotes the generator loss of patches $P_\text{PAT},P_\text{HS}$, and $P_\text{VS}$, respectively. $\mathcal{L}_{\text{D}}(P_\text{intra}, \psi_\text{D})$ denotes the discriminator loss of negative samples $P_\text{intra}$, and $\mathcal{L}_{\text{D}}(\tilde{P}_\text{intra}, \psi_\text{D})$ denotes the discriminator loss of positive samples $\tilde{P}_\text{intra}$.

\noindent\textbf{Inter-Patch Smoothness Loss:}\quad To ensure the \textbf{smoothness and continuity} among neighboring patches, we additionally randomly crop $M_\text{inter}$ patches $P_\text{inter} = \{p^m_\text{inter} |\ m = 1, 2, \cdots, M_\text{inter}\}$ of size $h\times w$ from the generated attack pattern $\hat{P}^{\text{raw}}$ and treat them as negative samples, where $M_\text{inter}$ is a hyper-parameter. To balance between the positive and negative samples, we then randomly crop the same number of patches $\tilde{P}_\text{inter} = \{\tilde{p}^m_\text{inter} |\ m = 1, 2, \cdots, M_\text{inter}\}$ from the ground truth pattern $\tilde{P}$.
We feed $P^{\text{inter}}$ and $\tilde{P}^{\text{inter}}$ to $\text{D}$ and compute the loss as:
\begin{equation}
\small
    \mathcal{L}_\text{inter} = \mathcal{L}_{\text{D}}(P_{\text{inter}}, \psi_\text{D}) + \mathcal{L}_{\text{D}}(\tilde{P}_{\text{inter}}, \psi_\text{D})\\
\end{equation}
where $\mathcal{L}_{\text{D}}(P^{\text{inter}}, \psi_\text{D})$ and $\mathcal{L}_{\text{D}}(\tilde{P}^{\text{inter}}, \psi_\text{D})$ denotes the discriminator loss obtained from negative samples $P^{\text{inter}}$ and positive samples $\tilde{P}^{\text{inter}}$, respectively. For the details of how the generator loss and discriminator loss respect to the negative samples and positive samples being calculated, please refer to Arjovsky \emph{et al}. \cite{arjovsky2017wasserstein}.

Consequently, the final loss for training PQ-GAN is obtained by combining Inter- and intra-Patch Smoothness losses as:
\begin{equation}
    \mathcal{L}_{\text{PQGAN}} =\mathcal{L}_{\text{intra}} + \mathcal{L}_\text{inter}
\end{equation}
The combined loss would enforce three generators to be jointly learned for generating a smooth and continuous global attack pattern of any scale.

\begin{table*}[!t]
\renewcommand{\arraystretch}{0.85}
\centering
    \setlength{\aboverulesep}{0pt}
    \setlength{\belowrulesep}{0pt}
\setlength{\tabcolsep}{0.95mm}{
\small
\begin{tabular}{l|c|c|ccccccc}
\toprule[1pt]
Model & ATK Region & Attack Methods & ResNet-18 & VGG-19 & ResNet-50 & Inception-V3 & MobileNet-V3  \\ 
\whline
 &  &  None & 67.3 & 68.3 & 74.0 & 71.3 & 66.1\\
\whline

& \multirow{2}*{Pixel-wise} &$\text{FGSM}_{\ \text{ICLR'2014} }$ \cite{goodfellow2014explaining}  & $3.8^{*}$(63.5$\downarrow$)    & 54.3 (14.0$\downarrow$)  & 56.4 (17.6$\downarrow$) & 59.2 (12.1$\downarrow$)  & 53.0 (13.1$\downarrow$) \\
&    &   $\text{C\&W}_{\ \text{IEEE SP'2017} }$ \cite{carlini2017towards}  & \textbf{0.0$^{*}$(67.3$\downarrow$)}     &55.8 (12.5$\downarrow$)  & 57.6 (16.4$\downarrow$) & 60.1 (11.2$\downarrow$) & 53.1 (13.0$\downarrow$) \\
             \cline{2-8}
& \multirow{2}*{Local} 
& $\text{Shadow ATK}_{\ \text{CVPR'2022} }$ \cite{zhong2022shadows} & $16.7^{*}$(50.6$\downarrow$)     & 56.4 (11.9$\downarrow$) & 56.9 (17.1$\downarrow$) & 57.3 (14.0$\downarrow$)  & 51.4 (14.7$\downarrow$) \\ 
&    &  $\text{ColorFool}_{\ \text{CVPR'2020} }$ \cite{shamsabadi2020colorfool} & $9.3^{*}$(58.0$\downarrow$)     & 55.2 (13.1$\downarrow$) & 56.4 (17.6$\downarrow$) & 60.1 (11.2$\downarrow$)  & 53.5 (12.6$\downarrow$) \\
\cline{2-8}
            
        & \multirow{6}*{Global}   
         & $\text{IadvHaze}_{\ \text{Arxiv'2021} }$ \cite{gao2021advhaze} & $ 21.4^{*}$(55.9$\downarrow$)   & 54.5 (13.8$\downarrow$)  & 56.0 (18.0$\downarrow$) & 58.4 (12.9$\downarrow$) & 48.3 (18.8$\downarrow$)\\
         ResNet-18    &    &  $\text{RA-AVA}_{\ \text{IJCAI'2021} }$ \cite{tian2021ava}&   $4.2^{*}$(63.1$\downarrow$)     & 55.5 (12.8$\downarrow$)  & 54.7 (19.3$\downarrow$) & 55.6 (15.7$\downarrow$) & 48.0 (18.1$\downarrow$)\\

\cline{3-8}
            &    & \cellcolor{lightgray!25}   \text{Rain Streaks} (Ours)  & \cellcolor{lightgray!25}$ 3.6^{*}$(63.7$\downarrow$)     &\cellcolor{lightgray!25} 52.2 (16.1$\downarrow$)  &\cellcolor{lightgray!25} 54.6 (19.4$\downarrow$) & \cellcolor{lightgray!25}54.3 (17.0$\downarrow$) & \cellcolor{lightgray!25}48.0 (18.1$\downarrow$) \\
           &    &  \cellcolor{lightgray!25}  \text{Lens Dirt} (Ours) &\cellcolor{lightgray!25} $ 4.3^{*}$(63.0$\downarrow$)  & \cellcolor{lightgray!25}50.0 (18.3$\downarrow$)   &\cellcolor{lightgray!25} 54.5 (19.5$\downarrow$) &\cellcolor{lightgray!25} 55.7 (15.6$\downarrow$) & \cellcolor{lightgray!25}48.3 (17.8$\downarrow$)\\

           &    & \cellcolor{lightgray!25}  \text{Snow Flakes} (Ours) &\cellcolor{lightgray!25} $ 2.5^{*}$(64.8$\downarrow$)  & \cellcolor{lightgray!25}51.0 (17.3$\downarrow$)    &\cellcolor{lightgray!25} \textbf{53.6} (\textbf{20.4$\downarrow$}) &\cellcolor{lightgray!25} 56.1 (15.2$\downarrow$) &\cellcolor{lightgray!25} 48.8 (17.3$\downarrow$)\\
          &    & \cellcolor{lightgray!25} \text{Rain Drops } (Ours) &\cellcolor{lightgray!25}$ 2.1^{*}$(65.2$\downarrow$)  &\cellcolor{lightgray!25} \textbf{49.1} (\textbf{19.2$\downarrow$}) &\cellcolor{lightgray!25} 54.0 (20.0$\downarrow$) &\cellcolor{lightgray!25} \textbf{53.8} (\textbf{17.5$\downarrow$}) &\cellcolor{lightgray!25} \textbf{46.5} (\textbf{19.6$\downarrow$}) \\
                                  \midrule[1pt]
            & \multirow{2}*{Pixel-wise}& $\text{FGSM}_{\ \text{ICLR'2014} }$ \cite{goodfellow2014explaining}  \ & 54.8 (12.5$\downarrow$)         & $ 5.2^{*}$ (63.1$\downarrow$)       & 57.1 (16.9$\downarrow$)         & 58.3 (13.0$\downarrow$)       & 53.0 (13.1$\downarrow$)         \\
             &    &  $\text{C\&W}_{\ \text{IEEE SP'2017} }$ \cite{carlini2017towards} & 55.6 (11.7$\downarrow$)         & \textbf{0.0$^{*}$ (68.3$\downarrow$)}            & 58.2 (15.8$\downarrow$)          & 58.4 (12.9$\downarrow$)       & 52.6 (13.5$\downarrow$)    \\
             \cline{2-8}
& \multirow{2}*{Local} 
& $\text{Shadow ATK}_{\ \text{CVPR'2022} }$ \cite{zhong2022shadows} & 54.6 (12.7$\downarrow$) & $14.8^{*}$(53.5$\downarrow$) & 55.2 (18.8$\downarrow$) & 57.4 (13.9$\downarrow$)  & 50.6 (15.5$\downarrow$) \\
  &    & $\text{ColorFool}_{\ \text{CVPR'2020} }$ \cite{shamsabadi2020colorfool}  & 53.4 (13.9$\downarrow$)         & $ 10.1 ^{*}$ (58.2$\downarrow$)        & 55.2 (18.8$\downarrow$)         & 58.1 (13.2$\downarrow$)       & 53.0 (13.1$\downarrow$)             \\
\cline{2-8}
         & \multirow{6}*{Global}    &  $\text{IadvHaze}_{\ \text{Arxiv'2021} }$ \cite{gao2021advhaze}  & 55.1 (12.2$\downarrow$)         & $ 25.1^{*}$ (43.2$\downarrow$)        & 55.8 (18.2$\downarrow$)          & 60.0 (11.3$\downarrow$)         & 48.2 (17.9$\downarrow$)\\
         VGG-19      &    & $\text{RA-AVA}_{\ \text{IJCAI'2021} }$ \cite{tian2021ava} & 51.3 (16.0$\downarrow$)         &$ 5.1 ^{*}$ (63.2$\downarrow$)        & 54.6  (19.4$\downarrow$)        & 55.6 (15.7$\downarrow$)       & 48.1 (18.0$\downarrow$)  \\

\cline{3-8}
             &    &  \cellcolor{lightgray!25}  \text{Rain Streaks} (Ours)  &  \cellcolor{lightgray!25}   50.5 (16.8$\downarrow$)         & \cellcolor{lightgray!25} $ 3.4^{*}$  (64.9$\downarrow$)         & \cellcolor{lightgray!25}  52.8 (21.2$\downarrow$)         & \cellcolor{lightgray!25}  55.1 (16.2$\downarrow$)       & \cellcolor{lightgray!25}  \textbf{47.4} (\textbf{18.7}$\downarrow$) \\
             &    &  \cellcolor{lightgray!25}  \text{Lens Dirt} (Ours) & \cellcolor{lightgray!25}  49.5 (17.8$\downarrow$)         & \cellcolor{lightgray!25} $ 4.3 ^{*}$ (64.0$\downarrow$)         & \cellcolor{lightgray!25}  53.1 (20.9$\downarrow$)         & \cellcolor{lightgray!25}  55.6 (15.7$\downarrow$)       & \cellcolor{lightgray!25}  48.3 (17.8$\downarrow$)         \\
             &    &  \cellcolor{lightgray!25} \text{Snow Flakes} (Ours)  & \cellcolor{lightgray!25}    \textbf{49.3} (\textbf{18.0}$\downarrow$)  & \cellcolor{lightgray!25}  $2.1^{*}$(66.2$\downarrow$)  & \cellcolor{lightgray!25}  52.5 (21.5$\downarrow$)         &  \cellcolor{lightgray!25} 54.0 (17.3$\downarrow$)          &  \cellcolor{lightgray!25} 47.4 (18.7$\downarrow$)\\
             &    &  \cellcolor{lightgray!25}  \text{Rain Drops } (Ours)    & \cellcolor{lightgray!25}   49.8 (17.5$\downarrow$)         & \cellcolor{lightgray!25} $ 2.7^{*}$ (65.6$\downarrow$)           & \cellcolor{lightgray!25}  \textbf{52.1} (\textbf{21.9}$\downarrow$) & \cellcolor{lightgray!25}  \textbf{53.0} (\textbf{18.3}$\downarrow$) &  \cellcolor{lightgray!25} 47.8 (18.3$\downarrow$)   \\

\bottomrule[1pt]
\end{tabular}
}
{\caption{Accuracy (\%) of five trained models on clean images and adversarial examples. $^{*}$ indicates the result of the white box attack; ATK Region stands for attack region. Our attack methods beat all unrestricted attack methods under the white-box scenario. Although the traditional $L_p$ restricted attack methods (C$\&$W and FGSM) have great performance in the white-box scenario, the attack strength cannot be transferred to the unseen models. Under the black-box scenario, our methods beat all other methods with great improvement, indicating that our attack method is more transferable across unseen models.} \label{tab:transferability} }
\vspace{-0.2cm}
\end{table*}

\begin{figure}[t]
\centering
\includegraphics[width=0.9\columnwidth]{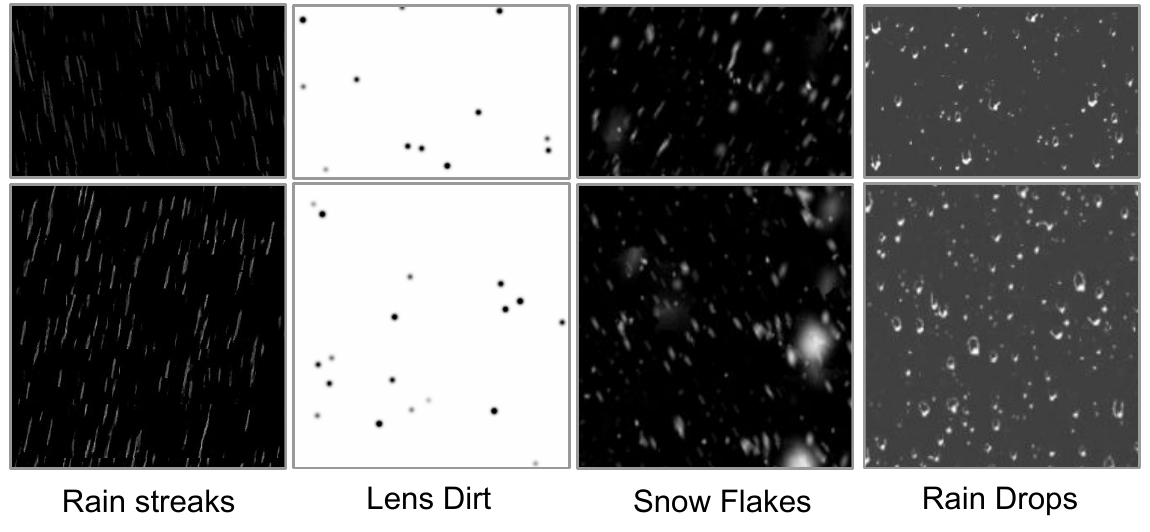}
    \caption{Four different patterns generated by our Patch Quilting Generator. PQG can generate patterns of any scale with great variety which is the key to the adversarial attack strength.}\label{fig:pattern}
\label{patterns}
\vspace{-0.2cm}
\end{figure}

\section{Experiment}

In this section, we demonstrate the effectiveness of the proposed method under various settings. The experimental setup is introduced in Sec.~\ref{setup}. To evaluate the task-agnostic property of our method, we compare it with existing approaches based on three computer vision tasks, \emph{i.e.}, image classification, object detection, and instance segmentation. Specifically, we evaluate the white-box attack strength and black-box transferability in Sec.~\ref{tranferability}, as well as the robustness against the defense algorithms in Sec.~\ref{sec:robustness}. We also compare the quality of the adversarial examples produced by our method and others. In addition, we adopt three no-reference image quality assessment metrics to quantify the image quality in Sec.~\ref{image_quality}. 

\subsection{Experimental Setup}
\label{setup}

\noindent \textbf{Dataset:}\quad The image classification experiments are conducted on 5000 randomly selected images from the validation set of the ImageNet \cite{imagenet_cvpr09} while the object detection and instance segmentation experiments are conducted on the entire validation set of the CityScapes dataset\cite{Cordts2016Cityscapes}.

\noindent \textbf{Implementation details:}\quad We employed 2000 $(256 \times 256)$ training samples for each of the four patterns: rain streak, rain drop, snow flakes, and camera lens dirt. Chen \emph{et al}.\cite{chen2021all} provides a set of snow flakes patterns where we randomly chose 2000 images for PQG training. Camera lens dirt patterns were generated by randomly adding 30 to 60 white points on a dark image and applying Gaussian blur with kernel standard deviation $\sigma=5$. Rain streak patterns are generated according to Garg \emph{et al.} \cite{garg2007vision}; rain drop pattern is an internet image and we perform Aguerrebere \emph{et al}. \cite{aguerrebere2013exemplar} to generate 2000 patterns of similar distribution. The synthesis strategy of rain drops, snow flakes, and lens dirt patterns are pixel-wise addition formulated as $I^{adv} = I + \gamma * P^I$. For rain streaks patterns, we performed depth-aware synthesis according to Hu \emph{et al}.\cite{hu2019depth}. Subsequently, we individually train four PQGs, each of which can generate various attack images containing the required pattern (See Fig.~\ref{patterns}). During the PQG-based Attack, each attack loop consists of 300 iterations with \cref{atk_iter} by default. Detailed parameter settings are provided in the supplementary material.

\begin{figure*}[t]
\centering
\includegraphics[width=1.85\columnwidth]{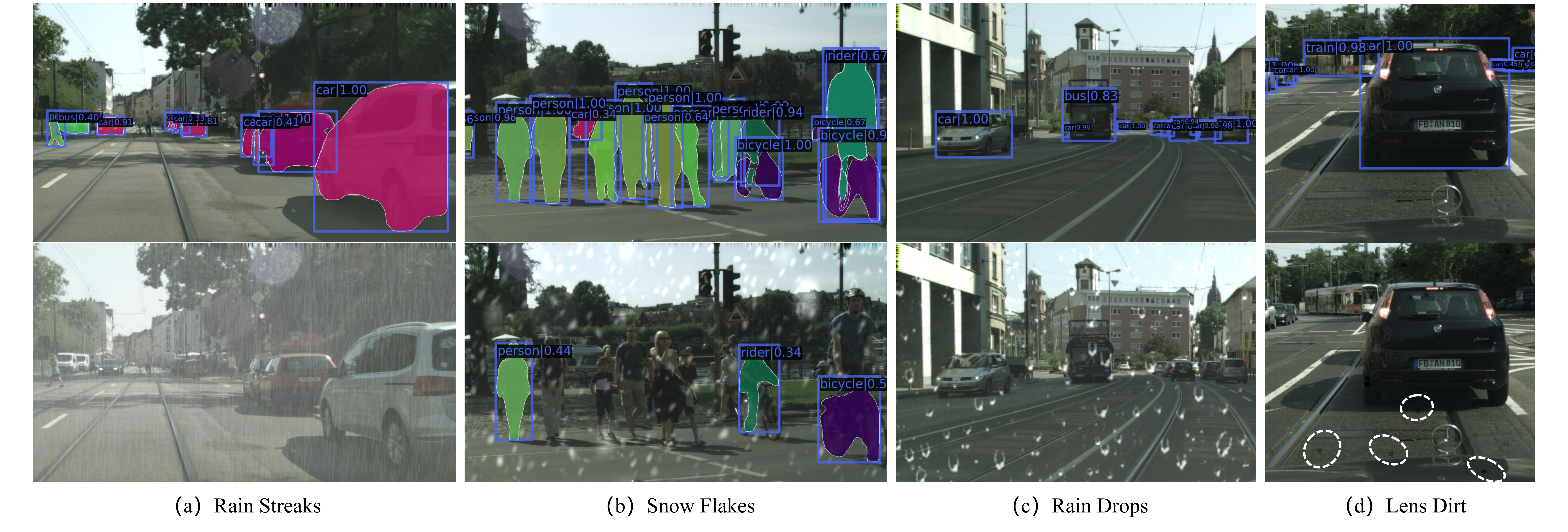}
    \caption{Four kinds of attack patterns generated by our approach. These patterns are scale-free, realistic, and misleading to instance segmentation models (the two columns on the left) and detectors (the two columns on the right).}
\label{ex_examples}
\vspace{-0.2cm}
\end{figure*}

\begin{table}[t]
\renewcommand{\arraystretch}{0.85}
\setlength{\tabcolsep}{2pt}
\setlength{\aboverulesep}{0pt}
\setlength{\belowrulesep}{0pt}
\begin{subtable}[b]{0.49\columnwidth}
\centering
\small
\setlength{\tabcolsep}{0.8mm}
\begin{tabular}{p{22mm}cc}
\toprule[1pt]
    \multirow{2}*{ATK Methods} & \multicolumn{2}{c}{mAP \% $\downarrow$}
    \\
    \cline{2-3} 
                                & Fr           & Mk\\ 
    \midrule[0.5pt]
     clean                      & 40.3              & 36.4\\
           \midrule[0.5pt]
    DPatch\cite{liu2018dpatch}  &8.8$^{*}$          &15.3 \\
    AdvPatch\cite{lee2019physical}&5.5$^{*}$        &9.6 \\
    UAP\cite{lee2019physical}&12.1$^{*}$        &12.2 \\
    \midrule[0.5pt]
   \rowcolor{mygray}RS (Ours)                &\textbf{4.2$^{*}$}        &\textbf{ 5.1} \\

                 \rowcolor{mygray}LD (Ours)      & 5.3$^{*}$         & 7.7 \\
              \rowcolor{mygray}SF (Ours)       & 4.9$^{*}$        & 4.8 \\
                 \rowcolor{mygray}RD (Ours)     & 5.8$^{*}$         & 5.5 \\
              \midrule[0.5pt]
\end{tabular}
\caption{\centering Object Detection (Target Network: Faster RCNN)}
\end{subtable}
\begin{subtable}[b]{0.49\columnwidth}
\centering
\small
\setlength{\tabcolsep}{0.8mm}
\begin{tabular}{p{22mm}cc}
\toprule[1pt]
    \multirow{2}*{ATK Methods} & \multicolumn{2}{c}{mAP \% $\downarrow$}
    \\
    \cline{2-3} 
    & Fr & Mk\\ 
      \midrule[0.5pt]
     clean & 40.3 & 36.4\\
          \midrule[0.5pt]
  DPatch\cite{liu2018dpatch}  &/         &/ \\
    AdvPatch\cite{lee2019physical}&/        &/ \\
    UAP\cite{mopuri2018generalizable}   &13.5 &6.3$^{*}$ \\
    \midrule[0.5pt]
    \rowcolor{mygray}RS (Ours)
     & 9.1 & \textbf{2.1$^{*}$} \\ 

    \rowcolor{mygray}LD (Ours)
     & 10.1 & 3.0$^{*}$ \\
    \rowcolor{mygray}SF (Ours)
     & \textbf{8.2} & 2.4$^{*}$ \\
    \rowcolor{mygray}RD (Ours)
     & 9.4 & 2.5$^{*}$ \\
              \midrule[0.5pt]
\end{tabular}
\caption{\centering Instance Segmentation (Target Network: Mask RCNN)}
\end{subtable}
\caption{Compare the cross-task transferability with other attack methods. RS, LD, SF, and RD are the abbreviations of our attack methods using Rain Streak, Lens Dirt, Snow Flakes, and Rain Drops patterns, respectively. Fr and Mk stand for Faster-RCNN and Mask-RCNN, respectively. Note that DPatch and AdvPatch are designed for attacking object detection models only.}
\label{cross-task}
\vspace{-0.8cm}
\end{table}

\begin{table}
\renewcommand{\arraystretch}{0.8}
\small
    \setlength{\tabcolsep}{0pt}
    \centering

    \begin{subtable}[ht]{1.0\linewidth}
    \centering
    
    \setlength{\aboverulesep}{0pt}
    \setlength{\belowrulesep}{0pt}
    \setlength{\tabcolsep}{0.8mm}{
        \begin{tabular}{p{20mm}|c|ccccc}
            \diagbox[width=21mm]{Attack}{Defense}  & None  & JPEG & HFC & HGD & APE& DEF-GAN  \\ 
 \midrule[1pt]
            FGSM\cite{goodfellow2014explaining}           & 3.8       & 61.5          & 64.7          & 64.7   & 61.9          &60.9\\
            C\&W\cite{carlini2017towards}         & \textbf{{0.0}}        & 64.0         & 65.5         &65.5         &64.8          &61.0  \\
            IadvHaze\cite{gao2021advhaze} &  21.4 & 45.5 & 46.3 & 35.7 & 41.4 & 35.7  \\
            RA-AVA\cite{tian2021ava} &  4.2& 40.2 & 41.6 & 43.9 & 48.9 & 42.2\\
            ColorFool\cite{shamsabadi2020colorfool} & 9.3 & 10.4 & 11.9 & 12.4 & 25.1 & 40.1 \\
            Shadow\cite{zhong2022shadows}&	16.7&	18.1&	18.5&	21.0&	22.7&	33.4 \\
             \midrule[1pt]
            \rowcolor{mygray} RS (Ours)& 3.6 & 12.1 & 11.5 & 13.2 & 15.9 & 26.5 \\
           \rowcolor{mygray} LD (Ours)&  4.3 & 10.3 & 12.8 & 12.2 & 21.9 & 28.0 \\
            \rowcolor{mygray} SF (Ours)&  2.5& 9.9 & 13.3 & 12.5 & \textbf{{14.5}}  & 23.4\\
            \rowcolor{mygray} RD (Ours)&  2.3&  \textbf{{9.7}} & \textbf{{10.7}} & \textbf{{12.1}} & 13.6  & \textbf{{25.8}}\\

        \end{tabular} \label{tab3-a}}
    \caption{Image Classification (ResNet-18).}
    \end{subtable}
    
        \begin{subtable}[ht]{1.0\linewidth}
        \centering
    \setlength{\aboverulesep}{0pt}
    \setlength{\belowrulesep}{0pt}
    \setlength{\tabcolsep}{0.8mm}{
        \begin{tabular}{p{20mm}|c|ccccc}
           \diagbox[width=21mm]{Attack}{Defense}  & None  & JPEG & HFC & HGD & APE & DEF-GAN  \\ 
 \midrule[1pt]
 DPatch\cite{liu2018dpatch} &8.8 &9.5 &10.6 &14.1 &13.9 &21.5 \\
 AdvPatch\cite{lee2019physical} &5.5 &7.1 &8.5 &10.5 &12.4 &15.7 \\
 UAP\cite{lee2019physical} &12.1 &14.7 &13.5 &15.2 &13.0 &13.2 \\
 \midrule[1pt]
        \rowcolor{mygray} RS (Ours)& \textbf{{4.2}}& 5.0 & 5.1 & \textbf{{5.6}} & 7.2 & \textbf{{8.8}} \\
        \rowcolor{mygray} LD (Ours)& 5.3 & \textbf{{5.9}} & \textbf{{5.6}} & 6.7 & 8.3 & 9.1 \\
        \rowcolor{mygray} SF (Ours)& 5.8 & 7.6 & 8.1 & 7.6 & \textbf{{7.9}} & 9.9 \\
        \rowcolor{mygray} RD (Ours)&  4.9& 6.4 & 7.1  & 8.2& 8.9 & 8.8\\

        \end{tabular}{\label{tab3-b} }}
    \caption{Object Detection (Faster-RCNN)}
    \end{subtable}
    
        \begin{subtable}[ht]{1.0\linewidth}
        \centering
        
            \setlength{\aboverulesep}{0pt}
    \setlength{\belowrulesep}{0pt}
    \setlength{\tabcolsep}{0.8mm}{
        \begin{tabular}{p{20mm}|c|ccccc}
            \diagbox[width=21mm]{Attack}{Defense}  & None  & JPEG & HFC & HGD & APE & DEF-GAN  \\ 
 \midrule[1pt]
  UAP\cite{mopuri2018generalizable} &6.3 &7.2 &9.4 &9.0 &8.6 &7.3 \\
   \midrule[1pt]
        \rowcolor{mygray} RS (Ours)& \textbf{{2.1}} & \textbf{{2.9}} & 4.4 & 4.7 & 5.4 & 6.0 \\
        \rowcolor{mygray} LD (Ours)& 3.0& 2.9 & \textbf{{4.3}} & \textbf{{4.2}} & 5.6 & 5.8  \\
        \rowcolor{mygray} SF (Ours)& 2.4& 2.9 & 4.3 & 4.6 & 5.2 & 6.0\\
        \rowcolor{mygray} RD (Ours)& 2.5 & 4.4 & 4.7 & 4.6 & \textbf{{4.7}} & \textbf{{5.2}} \\
        \end{tabular}{\label{tab3-c} }}
        \caption{Instance Segmentation (Mask-RCNN)}
    \end{subtable}
        \caption{Classification accuracy $\%$ (a) and mean average precision $\%$ (b,c) of the model on the adversarial examples generated by different attack methods (first columns) and the adversarial examples that different defense algorithms (first row) are applied.}
\label{robustness}
\vspace{-0.2cm}
\end{table}


\subsection{Transferability}
\label{tranferability}

\noindent \textbf{Cross-model Transferability:}\quad In our experiment, five classifiers including ResNet-18 \cite{he2016deep}, ResNet-50 \cite{he2016deep}, VGG-19 \cite{simonyan2014very}, Inception-V3 \cite{szegedy2016rethinking}, and MobileNet-V3 \cite{howard2019searching}, are employed. We also additionally compare our method with two traditional attack methods: FGSM\cite{goodfellow2014explaining} and C$\&$W\cite{carlini2017towards}, two unrestricted local attack methods: ColorFool\cite{shamsabadi2020colorfool} and Shadow Attack\cite{zhong2022shadows}, as well as two unrestricted global attack methods: Adversarial Vignetting Attack (AVA) \cite{tian2021ava} and Adversarial Haze\cite{gao2021advhaze}. We employ the classification accuracy as the evaluation metric, where lower classification accuracy indicates better attack performance. 

As shown in \cref{tab:transferability}, our methods achieved the best performance among all compared attack models. Compared to existing approaches, our method provides at least 17.80\% and 18.88\% average performance drops for ResNet-18 and VGG-19 based classifiers. For example, when our approach attacks ResNet-18 with the Rain Drops attack pattern and transfer the adversarial examples to VGG-19, the classification accuracy is largely decreased from 68.3\% to 49.1\%, while the second best only decreasing the corresponding system to 54.3\%. Fig. \ref{fig:comparison} also illustrates the adversarial examples produced from different methods on. It can be observed that our method can generate more photo-realistic attack patterns, but the other methods will destroy the structure (Shadow Atatck) or color (ColorFool) of the benign image.

\noindent \textbf{Cross-task Transferability:}\quad To further demonstrate the effectiveness of our approach, we then evaluate the transferability of the generated adversarial examples under the cross-task and cross-model scenarios, \emph{i.e.,} we generate the adversarial examples by the objection detection model (Faster RCNN \cite{ren2015faster}) and transfer to the instance segmentation model (Mask RCNN \cite{he2017mask}), and vice versa. we compare our results with a task-agnostic attack methods: UAP\cite{mopuri2018generalizable} and two task-specific attack methods: DPatch\cite{liu2018dpatch} and AdvPatch\cite{lee2019physical}. We employ the mean average precision (mAP \%) as the evaluation metric, where lower mAP indicates better attack performance.

As shown in \cref{cross-task} (a), we employ the Faster RCNN as the surrogate model to generate the adversarial examples and test in the Mask RCNN. The results indicate that our attack decreases the original mAP by a large margin, \emph{i.e.}, 31.3\% with the Rain Streaks attack pattern, 30.9\% with the Rain Drops attack pattern, and so on. Compared with the current state-of-the-art method AdvPatch, we also achieve competitive results: the result of AdvPatch is 9.6\%, and our best result is 5.1\%, which is significantly lower. In \cref{cross-task}(b), we generate the adversarial patterns through Mask RCNN against the Fast RCNN. Our PQAttack still achieves competitive results. Using the Snow Flakes pattern, an improvement of 5.3\% is achieved by PQAttack (8.2\%) compared with UAP (13.5\%).

\subsection{Robustness}
\label{sec:robustness}
\begin{figure}[htbp]
    \centering
     \includegraphics[width=0.9\columnwidth]{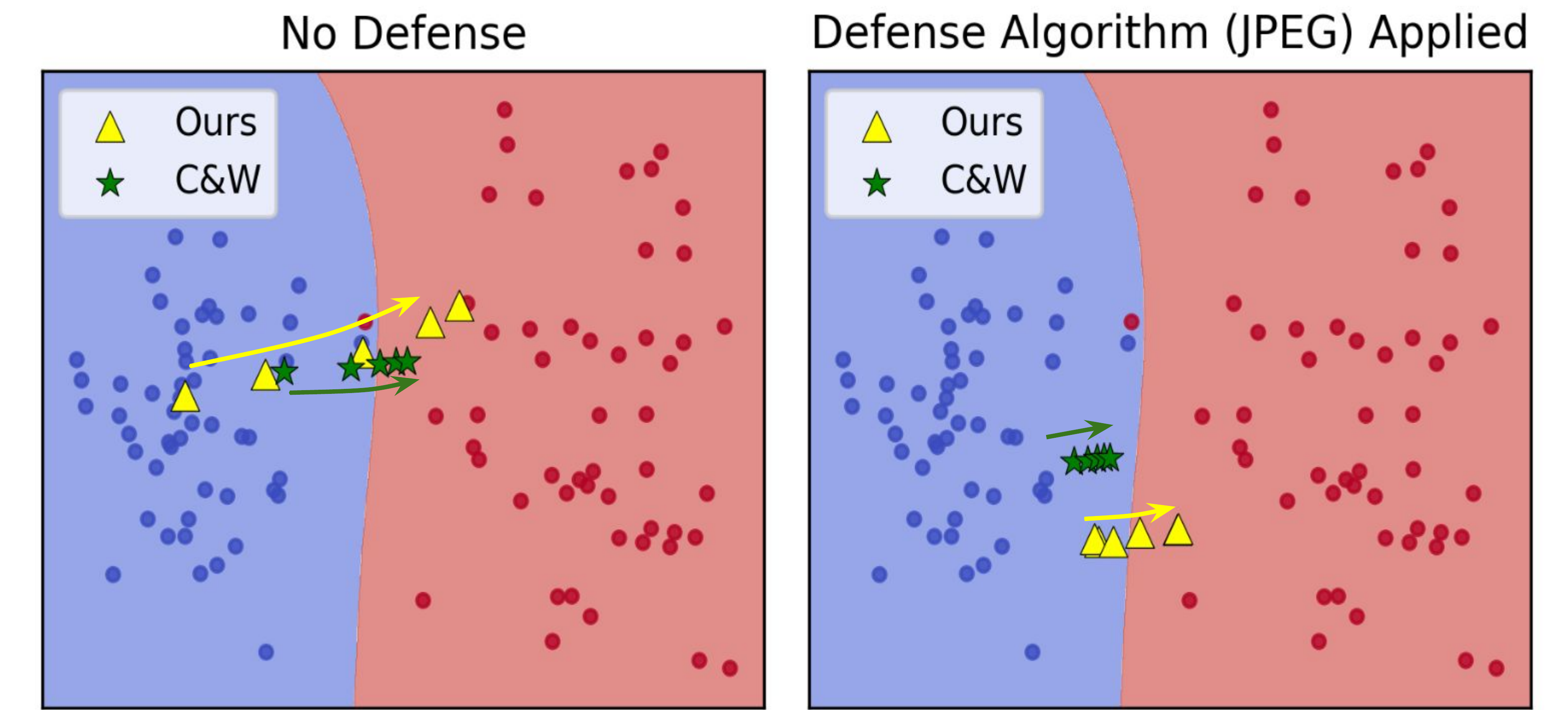}
     \caption{Adversarial trajectories visualized by the PCA\cite{abdi2010principal} dimension reduction algorithm. The decision boundaries are plotted by applying SVM\cite{hearst1998support} to the reduced vectors. The adversarial example gradually departs from its original class (blue) and moves to the target class (red) through the adversarial attack iterations. The JPEG compression defense algorithm\cite{das2018shield} successfully defense the restricted method C\&W (green stars) by pulling the adversarial examples back from the decision boundary, but it fails to defend our method because our method has no pixel-value restriction, and our adversarial example (yellow triangles) move much further apart from the decision boundary.}
    \label{atk_traj}
    \vspace{-0.2cm}
 \end{figure}

To evaluate the robustness, we apply three smoothing-based defense algorithms, JPEG Compression\cite{das2018shield}, High Frequency Suppression\cite{zhang2019adversarial}, HGD\cite{liao2018defense}, and two GAN-based defense algorithms Ape-GAN\cite{shen2017ape}, Defense-GAN\cite{samangouei2018defense} to the adversarial examples generated by our methods. Tab.~\ref{robustness} reports the experimental results on: (a) attacking image classification, (b) object detection, and (c) instance segmentation models. Smoothing-based defense algorithms (JPEG, HFC, HGD) are very effective in defending $L_p$-restricted attack methods (C$\&$W, FGSM, and UAP) but less effective in defending unrestricted attack methods. To illustrate this, we also visualize the adversarial trajectories in Fig.~\ref{atk_traj}.

Defense GAN\cite{samangouei2018defense} defense by learning the distribution of the benign images, which can identify and remove the disconformities between the distribution of the adversarial examples and benign images\cite{samangouei2018defense}, therefore very effective in defending the local attack methods (ColorFool\cite{shamsabadi2020colorfool}, Shadow Attack\cite{zhong2022shadows}, DPatch\cite{liu2018dpatch}, AdvPatch\cite{lee2019physical}). Different from the local attack methods, our method synthesizes semantic-aware adversarial patterns globally, which is very difficult to remove.


\subsection{Image Quality}
\label{image_quality}

For visualization, some typical adversarial examples of the Cityscapes and ImageNet are illustrated in Fig.~\ref{img_intro}, \ref{fig:comparison}, and \ref{ex_examples}. We observe that our method can generate patterns that are noticeable but natural to human eyes, \emph{i.e.}, it is hard for humans to identify the adversarial examples without referring to the original images. Meanwhile, the adversarial example can mislead the networks to give an incorrect output. It means that our generated semantic attack not only deceives the human visual system but also cheats the machine vision system.
\par We further employ three reference-free image quality assessment metrics: BRISQUE\cite{mittal2012no}, NIQE\cite{mittal2012making} and PIQE\cite{venkatanath2015blind} to quantify the image quality. We compare the average scores of the adversarial examples generated by different attack methods. The results shown in Tab.~\ref{numerical_assessment} demonstrate that the adversarial examples produced by our approach have the best image quality among all adversarial examples.


\begin{table}[t]
\small
\renewcommand{\arraystretch}{0.85}
\setlength{\tabcolsep}{0pt}
\centering
    \setlength{\aboverulesep}{0pt}
    \setlength{\belowrulesep}{0pt}
\setlength{\tabcolsep}{2.0mm}{
\begin{tabular}{p{22mm}|ccc}
\toprule[1pt]
ATK  & BRISQUE $\downarrow$ & NIQE $\downarrow$ & PIQE $\downarrow$\\
\midrule[0.5pt]
Clean & 22.4207 & 3.6792 & 32.1288\\
             \midrule[0.5pt]
            C\&W\cite{carlini2017towards}  & 32.1891 & 8.5136 & 36.1667\\
            FGSM\cite{goodfellow2014explaining} & 43.9118 & 17.8617 & 59.1998       \\
            IadvHaze\cite{gao2021advhaze} & 41.6842 & 12.7820 & 72.2287\\
            RA-AVA\cite{tian2021ava}  & 33.2389 & 9.3145 & 32.8531\\
            ColorFool\cite{shamsabadi2020colorfool} & 29.4896 & 5.1896 & 31.3078\\
            Shadow\cite{zhong2022shadows} &30.7310 & 5.1603 & 30.9756 \\ 
             \midrule[0.5pt]
 \rowcolor{mygray}           RS (Ours) & 30.3656 & 5.6763 & \textbf{30.5022}\\
 \rowcolor{mygray}           LD (Ours) & \textbf{27.3893} & 6.2135 & 31.8128\\
  \rowcolor{mygray}           SF (Ours) & 32.6419 & \textbf{4.8950} & 33.2251\\
  \rowcolor{mygray}           RD (Ours) & 31.2009 & 6.5481 & 32.9371\\

  \bottomrule[1pt]
                                  
\end{tabular}}
{\caption{Results of three non-reference image quality assessment metrics BRISQUE\cite{mittal2012no}, NIQE\cite{mittal2012making} and PIQE\cite{venkatanath2015blind} being evaluated on the adversarial examples generated by seven attack approaches. All adversarial examples are generated from the pre-selected 5000 images from the ImageNet dataset \cite{imagenet_cvpr09}.} \label{numerical_assessment}}
 \vspace{-0.2cm}
\end{table}

\section{Conclusion}

This paper proposes a novel PQ-GAN for adversarial attack, which learns a set of generators called PQG. This is the first generator-based approach that can generate \textbf{photo-realistic patterns of any scale}, which can be used to \textbf{attack various computer vision tasks}. The results show that our PQAttack approach achieved state-of-the-art results in misleading various white-box and black-box computer vision models. Importantly, the adversarial examples produced by our approach are not only robust to various defense algorithms but also have high visual qualities.

\noindent \textbf{Limitations and future works:}\quad The main limitation of our method is that our PQG can only generate limited types of attack patterns (e.g., rain and snow). However, it cannot generate patterns such as landscapes. we aim to address this in the future to generate more diversified scale-free attack patterns.

\noindent \textbf{Acknowledgements}\quad The work was supported by Natural Science Foundation of China under grants no. 62276170, 91959108, the Science and Technology Project of Guangdong Province under grant no. 2020A1515010707, the Science and Technology Innovation Commission of Shenzhen under grant no. JCYJ20190808165203670. The work of Siyang Song is funded by the EPSRC/UKRI under grant ref. EP/R030782/1.

{\small
\bibliographystyle{ieee_fullname}
\bibliography{arxiv_version}
}

\setcounter{figure}{0}  
\setcounter{table}{0}  
\renewcommand\thefigure{S\arabic{figure}}
\renewcommand\thetable{S\arabic{table}}

\appendix

\section{Results of Cross-model Transferability}

We additionally evaluate the cross-model transferability of our attack method on the Object Detection and Instance Segmentation task. For Object Detection, we choose Faster-RCNN\cite{ren2015faster} to be the target networks and transfer the adversarial examples to YOLOv3\cite{redmon2018yolov3} and Deformable DETR\cite{zhu2020deformable}. For Instance Segmentation, we choose Mask-RCNN\cite{he2017mask} to be the target networks and transfer the adversarial examples to PointRend\cite{kirillov2020pointrend} and SOLO\cite{wang2020solo}. The results are shown on Tab.~\ref{cross-model-transferability}.

\begin{table}[h]
\renewcommand{\arraystretch}{0.85}
\small
    \setlength{\tabcolsep}{0pt}
    \centering

    \begin{subtable}[h]{1.0\linewidth}
    \centering
    
    \setlength{\aboverulesep}{0pt}
    \setlength{\belowrulesep}{0pt}
    \setlength{\tabcolsep}{0.8mm}{
        \begin{tabular}{lccc}
        
            \midrule[1pt]
            \multicolumn{1}{c}{\multirow{2}{*}{Attack Methods}} & \multicolumn{3}{c}{mAP \%$\downarrow$}                \\
            \cline{2-4} 
            \multicolumn{1}{c}{}                             & FR-RCNN\cite{ren2015faster} & YOLO\cite{redmon2018yolov3} & DETR\cite{zhu2020deformable} \\ \midrule[0.5pt]
            clean                                            & 40.3        & 32.8   & 46.6            \\
            \midrule[0.5pt]
            DPatch\cite{liu2018dpatch}                                           & 8.8$^*$        & 12.3   & 20.2            \\
            AdvPatch\cite{lee2019physical}                                         & 5.5$^*$         & 12.5   & 15.4            \\
            UAP\cite{mopuri2018generalizable}                                             & 12.1$^*$        & 11.5   & 13.5            \\
            \midrule[0.5pt]
\rowcolor{mygray} Rain Streaks (Ours)                       & \textbf{4.2$^*$}         & 7.8    & \textbf{9.2}             \\
\rowcolor{mygray} Lens Dirt (Ours)                          & 5.3$^*$         & 10.2   & 12.4            \\
\rowcolor{mygray} Snow Flakes (Ours)                        & 4.9$^*$         & 8.3    & 9.7             \\
\rowcolor{mygray} Rain Drops (Ours)                         & 5.8$^*$         & \textbf{7.7}    & 11.0           \\
\midrule[1pt]
        \end{tabular}}
    \caption{Object Detection (Target Network: Faster-RCNN)}
    \end{subtable}
    
    \begin{subtable}[ht]{1.0\linewidth}
        \centering
    \setlength{\aboverulesep}{0pt}
    \setlength{\belowrulesep}{0pt}
    \setlength{\tabcolsep}{0.8mm}{

    \begin{tabular}{lccc}
        \midrule[1pt]
        \multicolumn{1}{c}{\multirow{2}{*}{Attack Methods}} & \multicolumn{3}{c}{mAP \%$\downarrow$}      \\
        \cline{2-4} 
        \multicolumn{1}{c}{}                             & Mk-RCNN\cite{he2017mask} & PtRend\cite{kirillov2020pointrend} & SOLO\cite{wang2020solo} \\ \midrule[0.5pt]
        clean                                            & 36.4      & 37.1      & 34.9 \\
        \midrule[0.5pt]
        UAP\cite{mopuri2018generalizable}                                              & 6.3$^*$       & 9.7       & 8.5  \\
        \midrule[0.5pt]
        \rowcolor{mygray} Rain Streaks (Ours)                              & \textbf{2.1$^*$}       & 7.8       & \textbf{7.3}  \\
        \rowcolor{mygray} Lens Dirt (Ours)                                 & 3.0$^*$         & 10.1      & 9.4  \\
        \rowcolor{mygray} Snow Flakes (Ours)                               & 2.4$^*$       & \textbf{7.5}       & 8.3  \\
        \rowcolor{mygray} Rain Drops (Ours)                                & 2.5$^*$       & 8.5       & 8.8 \\
    \midrule[1pt]
    \end{tabular}}
    \caption{Instance Segmentation (Target Network: Mask-RCNN)}
    
    \end{subtable}
    \caption{The performances of cross-model transferability on the object detection and instance segmentation task with other attack methods. The white box attack results are marked with $*$.}
\vspace{-2mm}
\label{cross-model-transferability}
\end{table}


\section{Explainability and Sensitivity Analysis}

\subsection{PQ-GAN Architecture}\label{pqg_a}
\begin{figure}[ht]
\centering
\includegraphics[width=1\columnwidth]{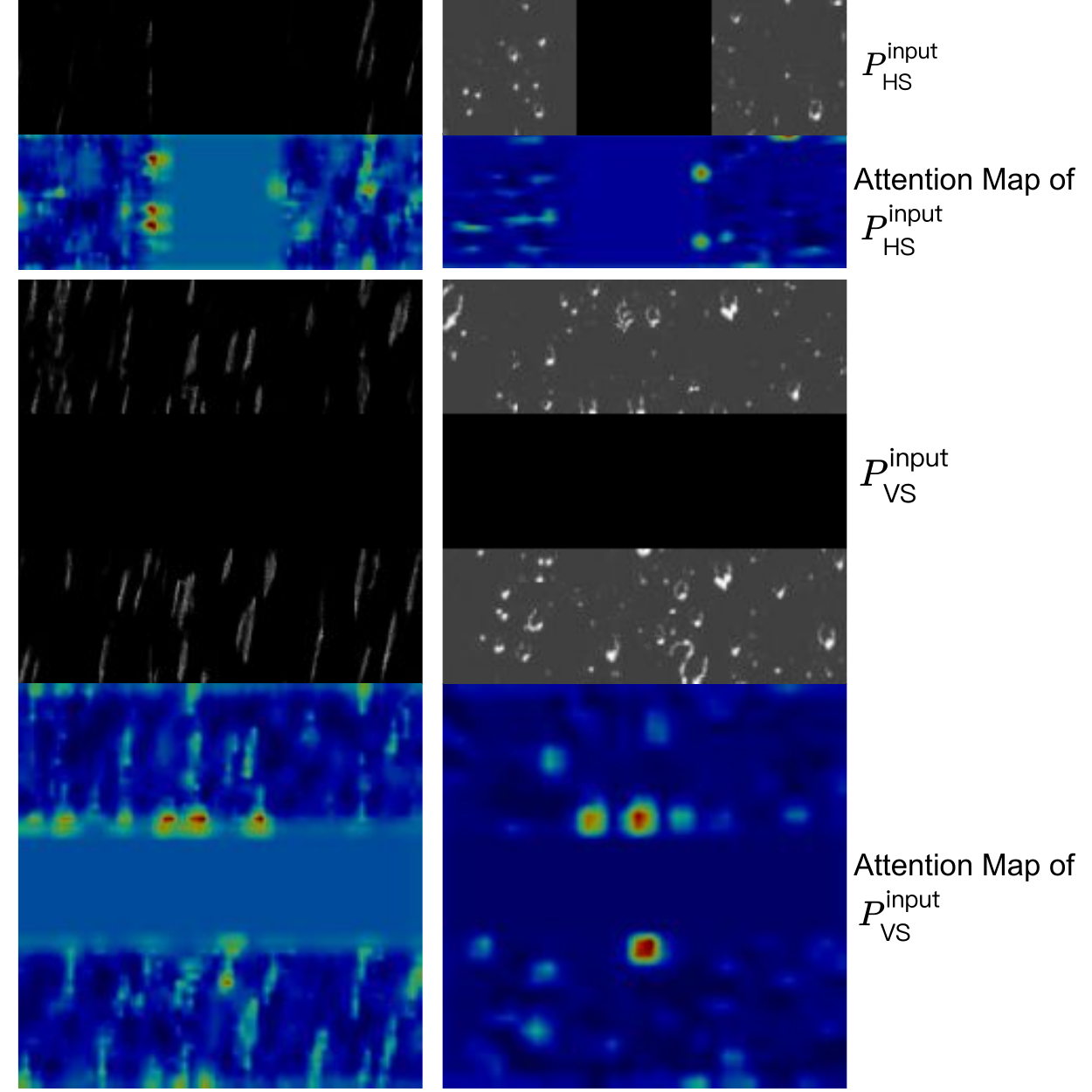} 
    \caption{Visualization of $P_\text{HS}^\text{input}$ and $P_\text{VS}^\text{input}$ with their attention map obtained by applying Ramaswamy \emph{et al.} \cite{ramaswamy2020ablation}. $\text{G}_\text{HS}$ and $\text{G}_\text{VS}$ pay the most attention to the edges between generated and empty patches, which results in the smoothness between neighbor patches.}
\label{Pinput}
\end{figure}

\begin{figure*}[t]
\setlength{\linewidth}{\textwidth}
\setlength{\hsize}{\textwidth}
\centering
\includegraphics[width=1.9\columnwidth]{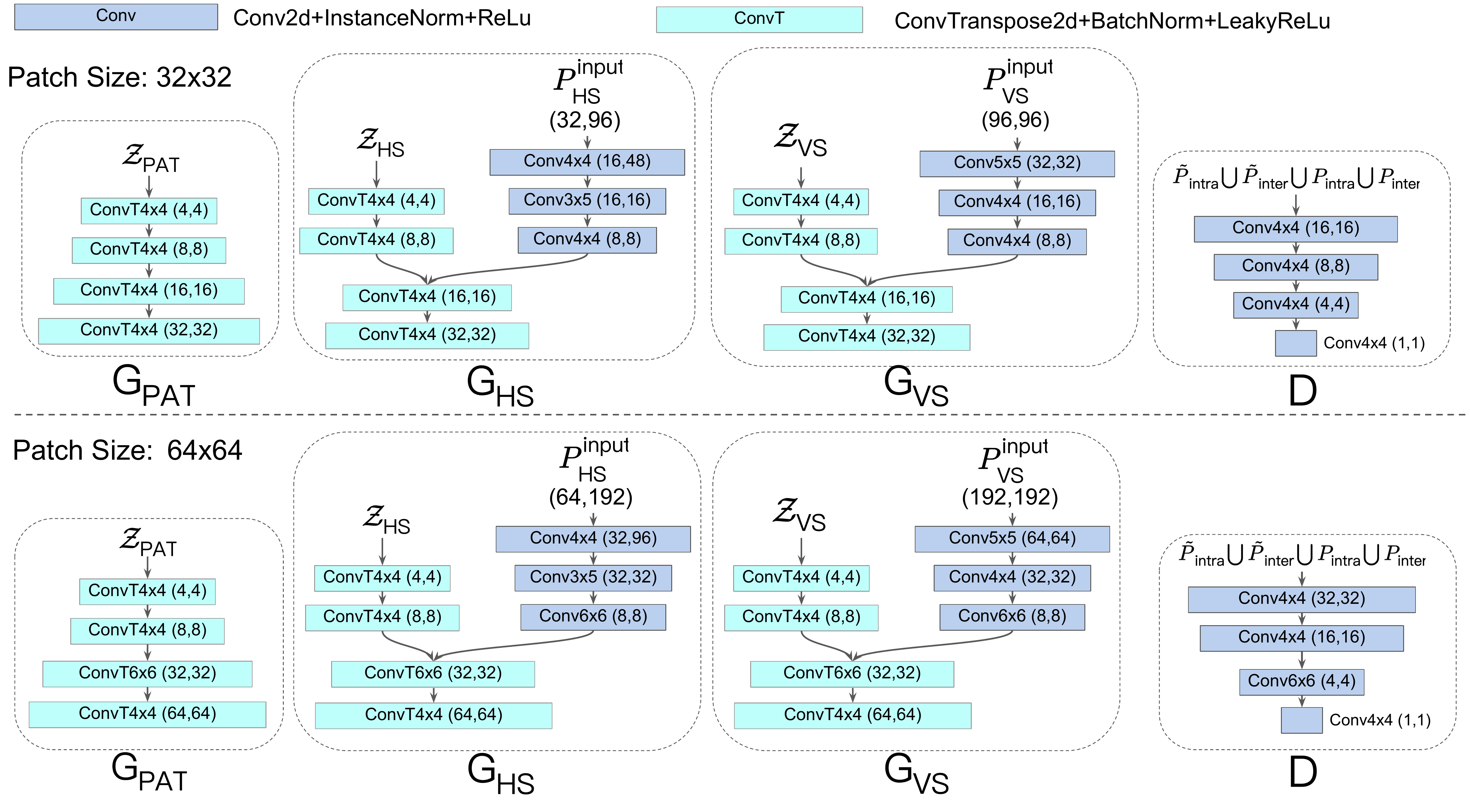}
\caption{Two PQ-GAN architecture of patch size (32x32) and (64x64).}
\label{Architecture}
\end{figure*}

To assist further analysis, we provide the details of the proposed PQ-GAN architecture in Fig.~\ref{Architecture}. In \emph{Eq}.~\ref{eqHS} and \ref{eqVS}, we introduce the input and output of the generators $\text{G}_\text{HS}$ and $\text{G}_\text{VS}$. To better extract the spacial relation between patches, we concatenate $p^\text{raw}_{2a\pm1,2b+1}$ and a zero patch of the same size to get $P_\text{HS}^\text{input}$ of scale $h \times 3w$ to be the input of $\text{G}_\text{HS}$, where $h,w$ is the pre-defined patch size. Similarly, we concatenate $p^\text{raw}_{2a\pm1,2b\pm1}$, $p^\text{raw}_{2a,2b\pm1}$ and three zero patches to get $P_\text{VS}^\text{input}$ of scale $3h\times 3w$ to be the input of $\text{G}_\text{VS}$, by Examples of $P_\text{HS}^\text{input}$ and $P_\text{VS}^\text{input}$ are depicted in Fig.~\ref{Pinput}.

\subsection{Explainability of the Patch-wise Smoothness}
$\text{G}_\text{HS}$ and $\text{G}_\text{VS}$ generate horizontal smoothness patches and vertical smoothness patches by considering their neighbor patches, $i.e.$, each $p^{\text{raw}}_{2a-1,2b}\in P_\text{HS}$ is generated conditioned to $P_\text{HS}^\text{input}$ which contains $p^\text{raw}_{2a\pm1,2b+1}$, while each $p^{\text{raw}}_{2a,2}\in P_\text{VS}$ is conditioned to $P_\text{VS}^\text{input}$ which contains $p^\text{raw}_{2a\pm1,2b\pm1}$ and $p^\text{raw}_{2a,2b\pm1}$. We draw the attention map of $P_\text{HS}^\text{input}$ and $P_\text{VS}^\text{input}$ to show how $\text{G}_\text{HS}$ and $\text{G}_\text{VS}$ pay attention to the neighbor patches, which is shown in Fig.~\ref{Pinput}.

\subsection{Influence of the Latent Vector's Dimensions}
\begin{figure}[htbp]
    \centering
    \includegraphics[width=1\columnwidth]{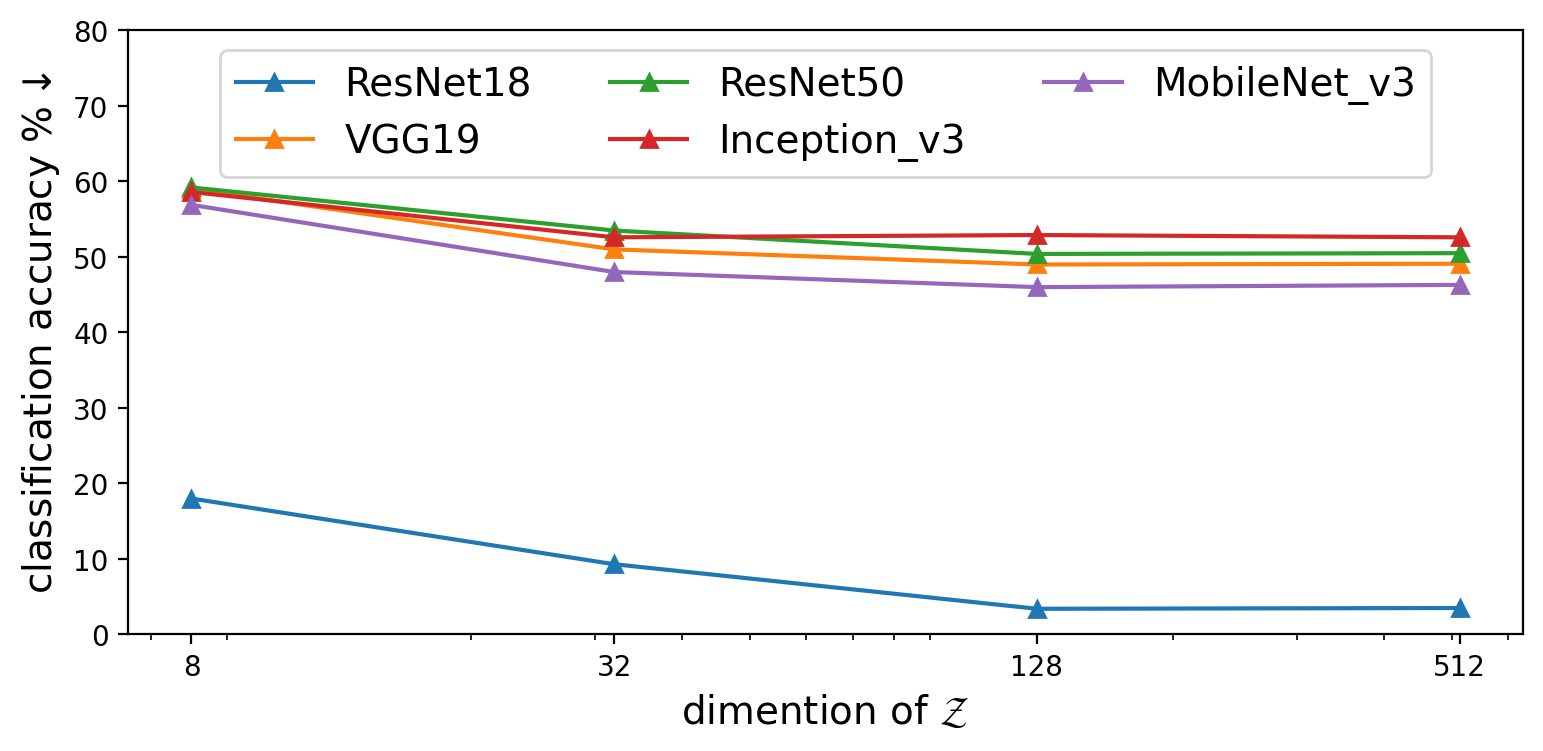}
    \caption{The white-box and black-box attack performance evaluation on the PQG
    \textit{w.r.t} to the different dimension (8, 32, 128 and 512) of the latent embeddings.}
    \label{diff_dim}
\vspace{-2mm}
\end{figure}

The adversarial attack strength is usually highly affected by the degree of freedom. For example, in traditional noise-based adversarial attack algorithms, tighter pixel-wise $l_p$ constraint usually leads to weaker attack strength. An extreme case is that one-pixel attack algorithm\cite{su2019one} has much lower attack strength than global attack algorithms. Instead of modifying the image pixel-wisely, our method modifies the latent embedding $\mathcal{Z}$ of PQG. We want to see how the dimension $k$ of the latent embedding $\mathcal{Z}$ affects the attack strength. We train PQG using Rain Streak samples with four different dimensions $k=\{8,32,128,512\}$ and use it to evaluate the white-box attack performance and black-box transferability on the ImageNet classification task. We can see from Fig.~\ref{diff_dim} that the classification accuracy decrease as the $k$ increases. It is especially influential under the white box scenario and when $k$ is small. The classification accuracy tends to be steady as $k$ goes above 128.

\section{Hyper-parameters Setups}\label{hyper_parameters}
\noindent \textbf{PQ-GAN:}\quad We conduct two different patch sizes, $32\times 32$ and $64\times 64$. We use patch size of $64\times 64$ for training the PQGAN of the rain streaks and snow flakes patterns and $32\times 32$ for training the PQGAN of the rain drops and lens dirt patterns. The dimension $k$ of each latent vector in $\mathcal{Z}$ is 128. During PQG training time, PQG generates images of size 256 $\times$ 256 with batch size set to be 4. The whole dataset (2000 images) is iterated 8 times. For the Inter-Patch Smoothness Loss, we set $M_\text{inter}=32$. Following the training setup of the Wasstarian GAN with gradient penalty\cite{gulrajani2017improved}, we use $\lambda=10,\ n_\text{critic}=5,\ \alpha=0.0001,\ \beta_1=0, \beta_2=0.9$, where $\lambda$ is the coefficient of the gradient penalty; $n_\text{critic}$ is the ratio of generator updates to each discriminator updates; $\alpha$ is the learning rate; $\beta_1$ and $\beta_2$ are the hyperparameters of the Adam\cite{kingma2014adam} optimizer. For more information, please refer to Gulrajani \emph{et al.} \cite{gulrajani2017improved}.

\noindent \textbf{PQAttack:}\quad The synthesis strategy of rain drops, snow flakes, and lens dirt patterns are pixel-wise addition are formulated as $I^{adv} = I + \gamma * P^I$, where $\gamma=0.3$ is used for rain drops and snow flakes patterns, $\gamma=-0.3$ is used for lens dirt patterns. For rain streaks patterns, we performed depth-aware synthesis according to Hu \emph{et al}.\cite{hu2019depth}, where we use $\alpha=0.03,\ \beta=0.04$. During the PQG-based Attack, each attack loop consists 300 iterations by default, where Adam \cite{kingma2014adam} optimizer is employed with $\beta_1=0.9$, $\beta_2=0.999$  and a starting learning rate $lr=0.03$. We also use cosine annealing scheduler \cite{loshchilov2016sgdr} for learning rate decay.

\noindent \textbf{Pre-trained Target Networks:} All image classification pre-trained weights are obtained from the PyTorch model library. Pretrained weights of Mask-RCNN and Faster-RCNN are obtained from the OpenMMLab. YOLOv3\cite{redmon2018yolov3}, Deformable BETR\cite{zhu2020deformable}, PointRend\cite{kirillov2020pointrend}, and SOLO\cite{wang2020solo} are trained using Cityscapes\cite{Cordts2016Cityscapes} standard training set with pre-trained weights on COCO dataset\cite{lin2014microsoft}.  Deformable BETR, PointRend, and SOLO are trained with batch size 8 for 64 epochs, while YOLOv3 are trained with batch size 32 for 273 epochs.

\noindent \textbf{Attack Methods:} We follow the standard hyper-parameter setup for all attack methods.

\clearpage
\section{Patterns generated by PQG}
\begin{figure*}[bp]
\centering
\includegraphics[width=1.95\columnwidth]{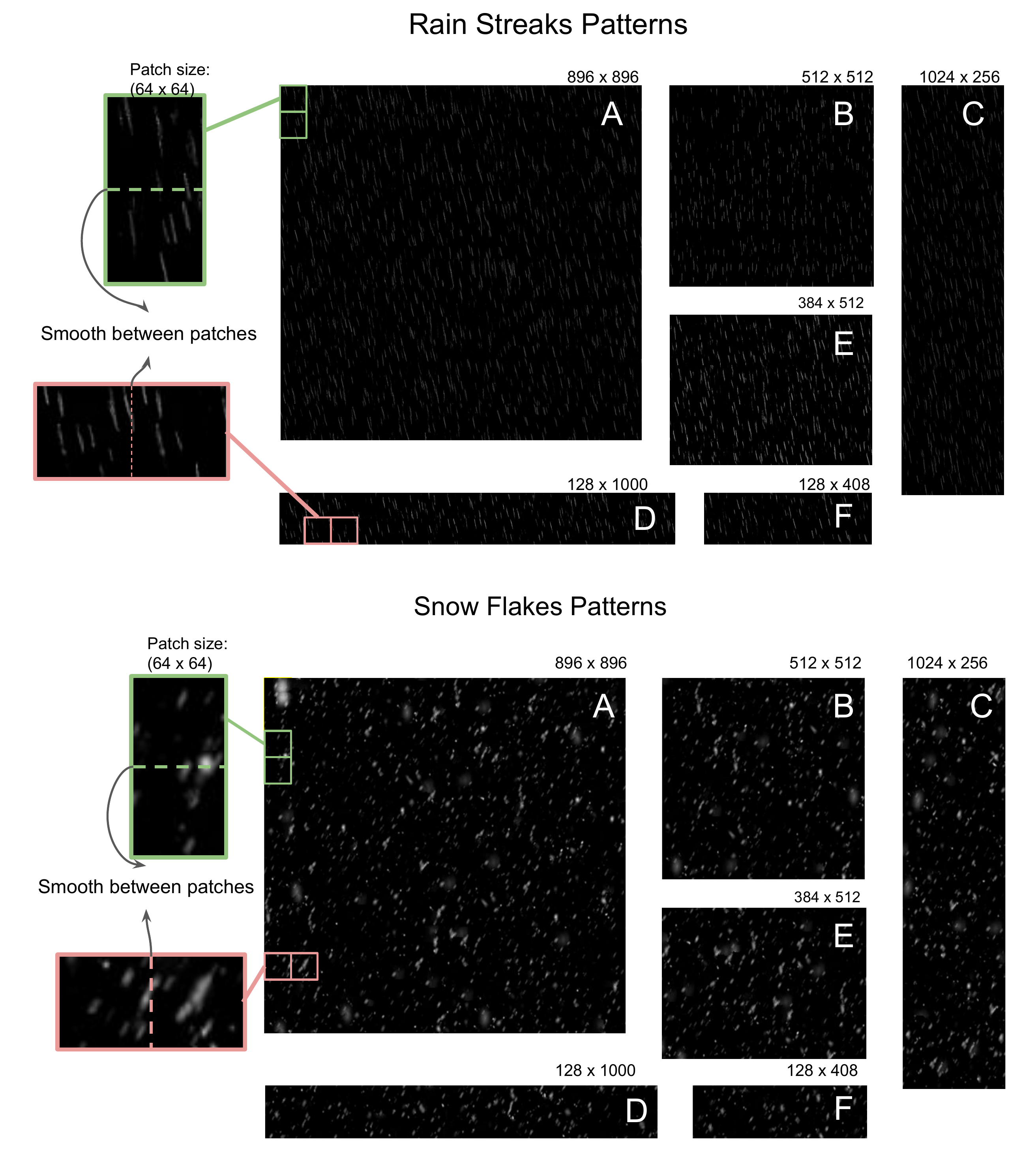}
\caption{Rain streaks and snow flakes patterns generated by our proposed PQG with patch size equals 64. Neighbor patches (marked in yellow and pink rectangles) are connected smoothly.}
\label{More patterns_1}
\end{figure*}

\begin{figure*}[b]
\centering
\includegraphics[width=2\columnwidth]{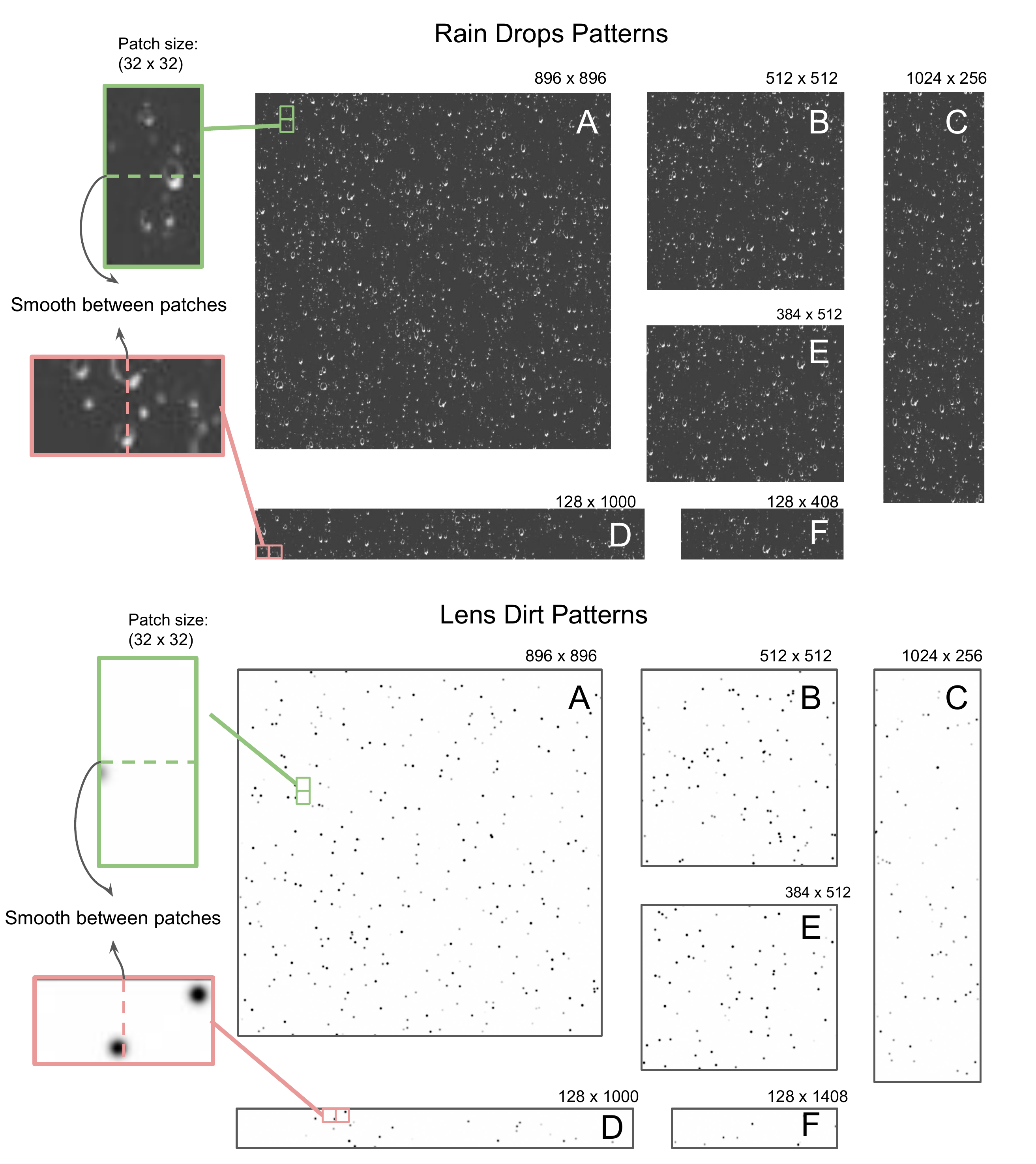}
\caption{Snow Flakes and lens dirt patterns generated by our proposed PQG with patch size equals 32. Neighbor patches (marked in yellow and pink rectangles) are connected smoothly.}
\label{More patterns_2}
\end{figure*}

Our proposed PQG can be used to generate patterns of any size without retraining the network. We display patterns of size 896$\times$ 896 (A), 512$\times$ 512 (B), 1024 $\times$ 256 (C), 128 $\times$ 1000 (D), 384 $\times$ 512 (E), 128 $\times$ 408 (F) in Fig.~\ref{More patterns_1} and~\ref{More patterns_2}.

\clearpage
\section{Visualization of Our Adversarial Examples}
\begin{figure*}[b]
\centering
\includegraphics[width=1.95\columnwidth]{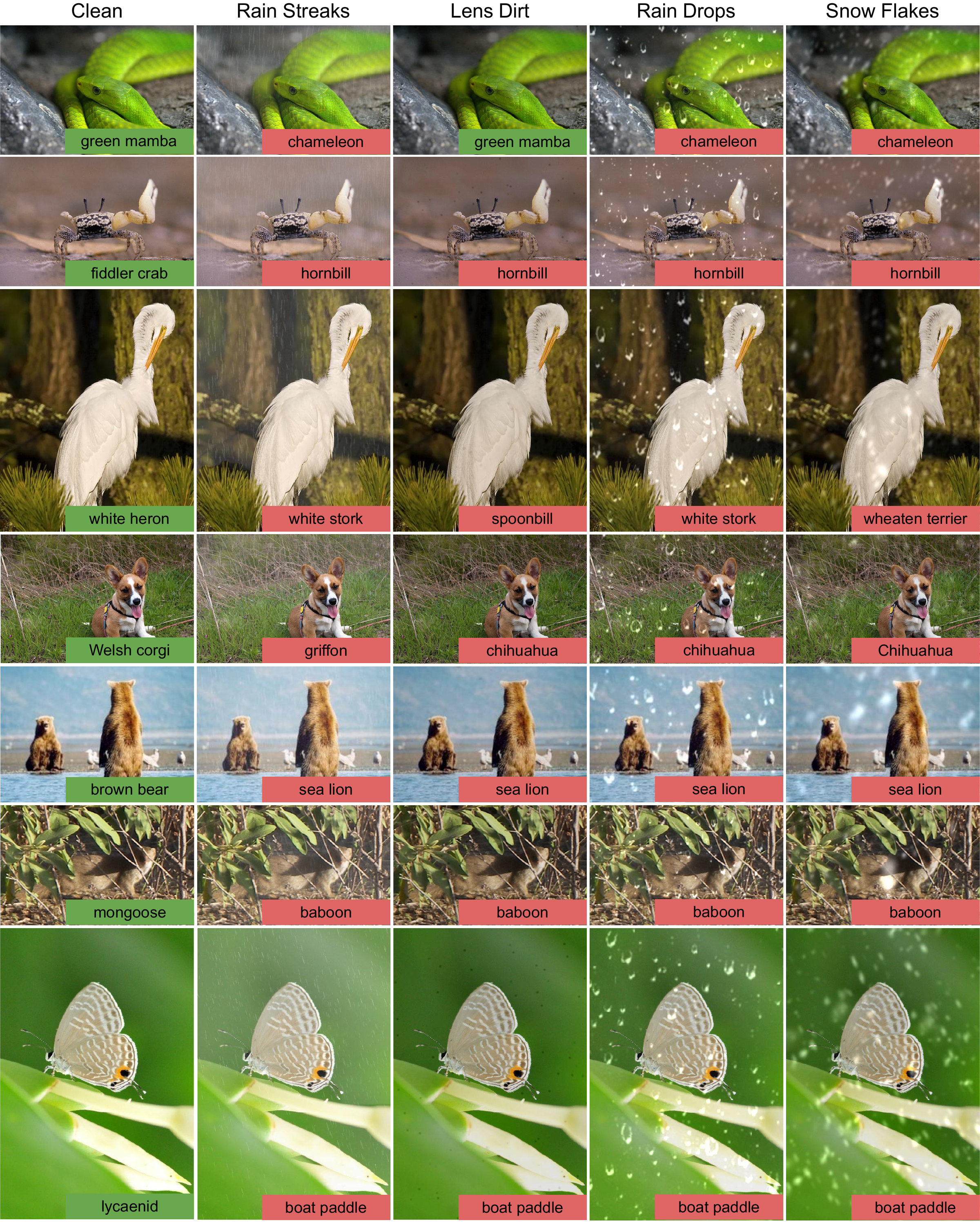}
\caption{Adversarial Examples of ImageNet dataset. The classification results are shown in the bottom right of each image.}
\label{more_visual_1}
\end{figure*}

\begin{figure*}[b]
\centering
\includegraphics[width=1.95\columnwidth]{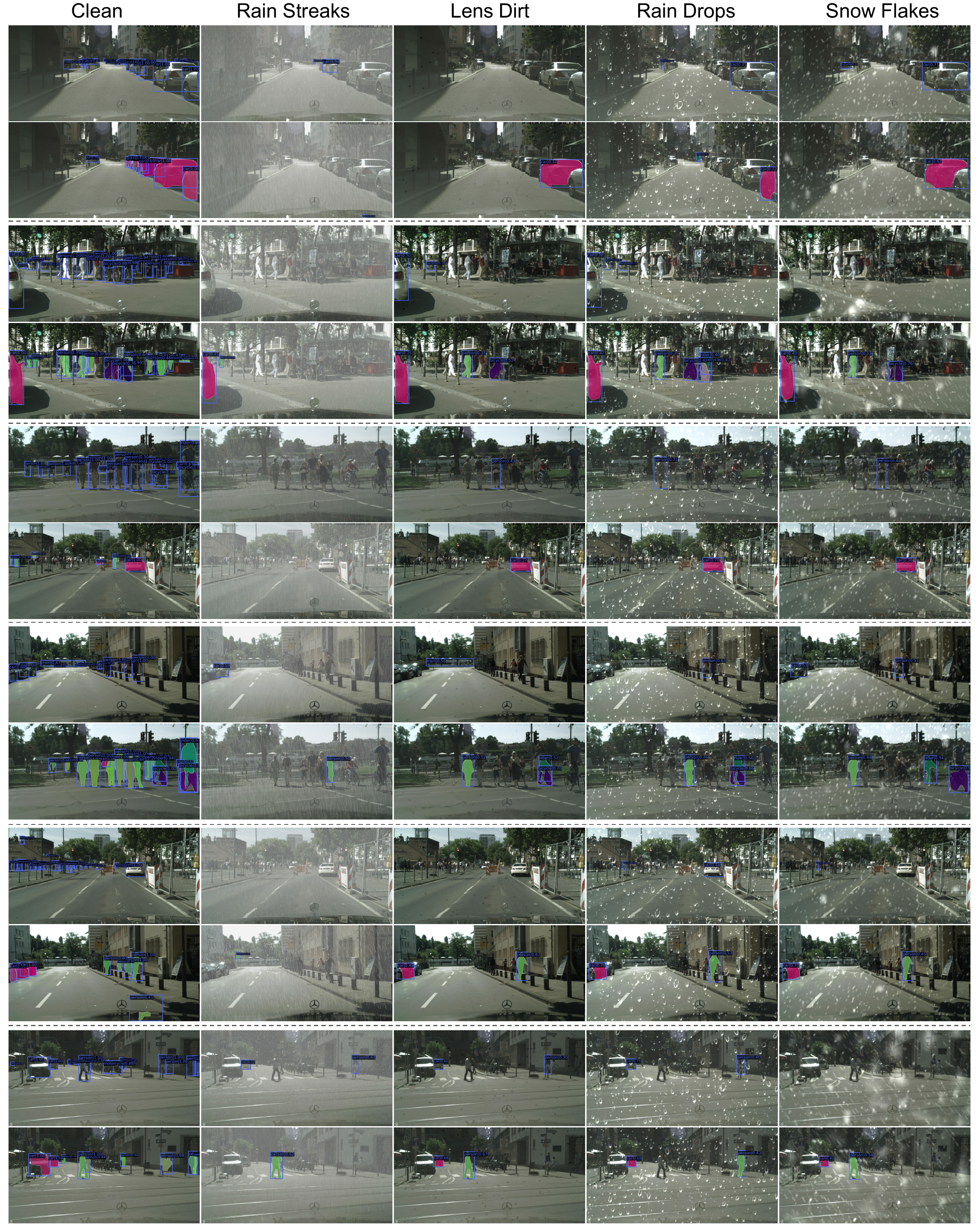}
\caption{Adversarial Examples of CityScapes dataset. The Object Detection and Instance Segmentation results are shown in the odd and even rows, respectively.}
\label{more_visual_2}
\end{figure*}
\end{document}